\documentclass[10pt,twocolumn,letterpaper,bold]{article}

\usepackage[pagenumbers]{cvpr} %

\definecolor{junglegreen}{rgb}{0.113, 0.639, 0.5}

\newcommand{\mhe}[1]

\newcommand{\methodterm}{RePerformer\xspace}

\definecolor{cvprblue}{rgb}{0.21,0.49,0.74}

\definecolor{custompink}{RGB}{255, 0, 144}  %

\newcommand{\myparagraph}[1]{\vspace{0.1em}\noindent\textbf{#1}}

\usepackage[pagebackref,breaklinks,colorlinks,allcolors=cvprblue]{hyperref}

\usepackage{multirow}
\def\paperID{5019} %
\def\confName{CVPR}
\def\confYear{2025}

 \title{RePerformer: Immersive Human-centric Volumetric Videos from Playback to Photoreal Reperformance}

\author{Yuheng Jiang$^{1,2*}$ \;\, Zhehao Shen$^{1,2*}$ \;\, Chengcheng Guo$^{1}$ \;\, Yu Hong$^{1}$ \;\, Zhuo Su$^{3}$ \;\, \\ Yingliang Zhang$^{4}$ \;\, Marc Habermann$^{5\dagger}$ \;\, Lan Xu$^{1\dagger}$} 

\makeatletter
\let\@oldmaketitle\@maketitle%
\renewcommand{\@maketitle}{
	\@oldmaketitle%
	\centering
	\vspace{-8mm}
	{\large \textsuperscript{1}ShanghaiTech University}\quad \quad
        {\large \textsuperscript{2}NeuDim}\quad \quad
	{\large \textsuperscript{3}ByteDance}\quad \quad
	{\large \textsuperscript{4}DGene}\quad \quad
    
        {\large \textsuperscript{5}Max Planck Institute for Informatics, Saarland Informatics Campus}\quad \quad

	\vspace{8mm}
}

\usepackage{colortbl}

\newcommand{\bestCellColor}[1]{\cellcolor[rgb]{.866,.945, 0.831}#1}

\newcommand{\secondBestCellColor}[1]{\cellcolor[rgb]{1, 0.98, 0.83}#1}

\begin{document}

\twocolumn[{
  \renewcommand\twocolumn[1][]{#1}
  \maketitle
  \begin{center}
  \vspace{-6ex}
  \includegraphics[width=1\textwidth]{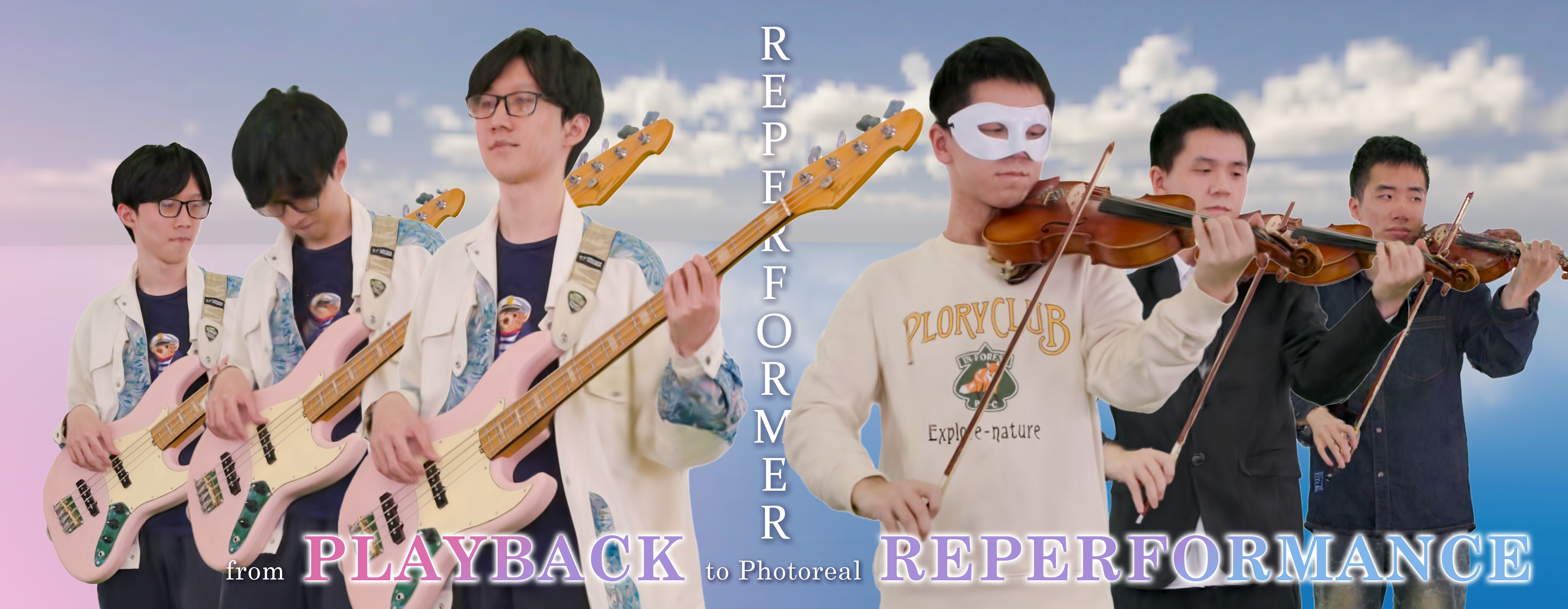}
  \vspace{-4ex}
  \captionof{figure}{We introduce \textbf{\methodterm}, a Gaussian-based approach for robust, high-fidelity volumetric video playback and realistic re-performance of general dynamic scenes. (Left: Sequential bass performances. Right: Synchronized violin players.)}
  \label{fig:teaser}
  \end{center}
}]

\def\thefootnote{*}\footnotetext{Equal Contribution. ${\dagger}$ Corresponding authors. }

\begin{abstract}
  
Human-centric volumetric videos offer immersive free-viewpoint experiences, yet existing methods focus either on replaying general dynamic scenes or animating human avatars, limiting their ability to re-perform general dynamic scenes. In this paper, we present RePerformer, a novel Gaussian-based representation that unifies playback and re-performance for high-fidelity human-centric volumetric videos. Specifically, we hierarchically disentangle the dynamic scenes into motion Gaussians and appearance Gaussians which are associated in the canonical space. We further employ a Morton-based parameterization to efficiently encode the appearance Gaussians into 2D position and attribute maps. For enhanced generalization, we adopt 2D CNNs to map position maps to attribute maps, which can be assembled into appearance Gaussians for high-fidelity rendering of the dynamic scenes. For re-performance, we develop a semantic-aware alignment module and apply deformation transfer on motion Gaussians, enabling photo-real rendering under novel motions. Extensive experiments validate the robustness and effectiveness of RePerformer, setting a new benchmark for playback-then-reperformance paradigm in human-centric volumetric videos. 
Project page: \href{https://moqiyinlun.github.io/Reperformer/}{\textcolor{custompink}{https://moqiyinlun.github.io/Reperformer/.}}

\end{abstract}

\section{Introduction}
As the boundary between digital and real worlds continues to blur, human-centric volumetric videos represent a significant advancement in visual media, allowing users to control the virtual camera view freely. This facilitates interactive exploration and various immersive applications in telepresence, education, and entertainment.

By far, the high-end solutions for volumetric videos utilize dense multi-view videos to faithfully record the dynamic scenes. Within this domain, we observe that there are mainly two complementary workhorses for producing such human-centric volumetric videos, dubbed \textit{``replay volumetric videos''} and \textit{``animate volumetric videos''} for short. The former playback setting~\cite{collet2015high,tretschk2021nonrigid,jiang2024robust} merely reconstructs and replays the dynamic scenes. In contrast, the latter animatable setting~\cite{habermann2021real,li2022tava, noguchi2021neural, uzolas2024template} can ``reperform'' the volumetric videos under novel motions unseen during recording. 
Both sides of research have witnessed significant changes in the underlying 3D representations, from early textured meshes~\cite{dou2016fusion4d,casas20144d, TotalCapture}, to NeRFs~\cite{zhao2022human,jiang2023instantavatar}, and to the very recent 3DGS~\cite{kerbl20233d,li2024gaussianbody,chen2024egoavatar}. Currently, in the playback setting, the Gaussian-based advances~\cite{liang2023gaufre, jiang2024robust} achieve high-fidelity rendering even for complex scenes like cooking or piano playing. Yet, such playback methods inherently lack the generalization ability to new motions. On the other hand, the animatable advances~\cite{li2024gaussianbody,chen2024egoavatar, zielonka25dega} mainly focus on human-only scenes (known as human avatars), heavily relying on human skeletons or parametric models like SMPL~\cite{SMPL2015}. They also emphasize the generation ability to arbitrary novel motions and hence require extensive motion sequences for training.

In this paper, we explore a novel direction for human-centric volumetric video, at the intersection of the above traditional playback and animatable settings. Given a dynamic sequence with dense footage, our goal is not only deliver accurate free-view playback but also to realistically re-perform the dynamic scene under similar yet novel motions.
Such a novel playback-then-reperformance setting is highly valuable with broad applications. For example, one can transfer a professional violinist's motion to an unskilled actor in film production, resulting in realistic performances and reduced actor training costs. However, it remains challenging methodologically. Ideally, it requires animatable ability for general non-rigid scenes only given limited motion examples from a dynamic sequence, let alone maintaining photo-realism for, both, playback and re-performance.

To this end, we introduce \textit{RePerformer}, a novel 3DGS-based approach for generating human-centric volumetric videos from dense multi-view inputs. In stark contrast to prior works, it provides realistic playback and can vividly reperform the non-rigid scenes driven by similar yet unseen motions (see Fig.~\ref{fig:teaser}).
At the core of our RePerformer is a nuanced and hierarchical disentanglement of a dynamic Gaussian representation. We disentangle the general non-rigid scene into compact motion Gaussians and dense appearance Gaussians, akin to the Embedded deformation (ED) graph and mesh tracking~\cite{sumner2007embedded, li2008global}. We further disentangle the dense appearance Gaussians into structured position maps and Gaussian attribute maps, which are associated with generalizable neural networks.

More specifically, it is a trilogy in our RePerformer. Given dense multi-view videos, \textbf{for the first tracking stage}, we disentangle motion and appearance Gaussians to provide topology-consistent tracking of the non-rigid scene. Note that both the compact motion and dense appearance Gaussians are initialized in the canonical frame and linked via nearest-neighbor search. We optimize only the positions and rotations of motion Gaussians,
using them to warp the dense Gaussians into each frame.
Since the appearance Gaussians are topology-consistent, we disentangle them into 3D positions and other attributes, which are repacked into 2D position/attribute maps using a Morton-based parameterization~\cite{morton1966computer} to preserve the spatial adjacency.
\textbf{For the second training stage}, we perform self-supervised learning on the captured multi-view videos to learn a generalizable mapping from position maps to attribute maps. We adopt the U-Net architecture~\cite{ronneberger2015u} with attention layers to enhance global coherence. 
Once trained, given the underlying motion Gaussians, we can obtain the dense position maps through non-rigid warping and then infer the corresponding attribute maps. Both maps are assembled into dense appearance Guassians for photo-real rendering of the dynamic scene.
\textbf{For further re-performance}, for new performer with novel motions, we adopt a semantic-aware alignment module to associate the motion Gaussians of the new performer with the appearance Gaussians of the original one. We then extend the deformation transfer scheme~\cite{sumner2004deformation} into motion Gaussians, to achieve photo-real rendering under the new motions.
To summarize, our main contributions include: 
\begin{itemize} 
	\setlength\itemsep{0em}

	\item We identify a new playback-then-reperformance setting for the human-centric volumetric videos.
    
	\item We present a Gaussian-based solution to achieve high-fidelity rendering of general dynamic scenes, for both playback and re-performance.
    
	\item We propose a hierarchical disentanglement scheme and a Morton-based parameterization for learning generalizable dynamic Gaussians.
	
	\item We design a semantic-aware motion transfer module to facilitate faithful re-performance across sequences.
\end{itemize}

\section{Related Work}

\noindent{\textbf{Non-rigid Reconstruction.}} 
In the field of Non-rigid registration and reconstruction, numerous methods~\cite{suo2021neuralhumanfvv, sun2021HOI-FVV, UnstructureLan, fridovich2023k, cao2023hexplane, wang2023videorf} have been proposed to address this challenging problem. Sumner~\etal~\cite{sumner2007embedded} introduces the classical embedded deformation graph for aligning two meshes, while SoG~\cite{StollHGST2011} utilizes a Gaussian model to represent human movement. HighQualityFVV~\cite{collet2015high} utilizes tracked mesh sequences with texture video to capture detailed human performance. Building on the pioneering work DynamicFusion~\cite{newcombe2015dynamicfusion}, a series of approaches~\cite{dou2016fusion4d, motion2fusion, KillingFusion2017cvpr, slavcheva2018sobolevfusion, robustfusion, su2022robustfusionPlus, jiang2022neuralhofusion} extend this technique to handle challenging scenarios. 
Recent neural advances~\cite{nerf}, bypass explicit reconstruction to conduct volume rendering at photo-realism. Based on it, dynamic NeRF variants~\cite{tretschk2021nonrigid, park2021nerfies, weng_humannerf_2022_cvpr} maintain a canonical space and leverage the photometric differences to learn the deformation. With advancements such as hash tables~\cite{muller2022instant} and tensor decomposition~\cite{chen2022tensorf}, several methods~\cite{isik2023humanrf, zhao2022human, jiang2023instant, song2023nerfplayer} significantly accelerate the training and rendering speed in the dynamic setting. However, rendering quality remains limited. The emergence of 3D Gaussian Splatting~\cite{kerbl20233d} marks a significant return to explicit modeling paradigms, with dynamic variants~\cite{luiten2023dynamic, li2023spacetime, duan20244d, guo2024motion, huang2024sc, labe2024dgd, gao2024gaussianflow, lin2024gaussian} advancing the field by tackling 4D scene reconstruction under complex scenarios. HiFi4G~\cite{jiang2024hifi4g} use mesh deformation to initialize Gaussians, while DualGS~\cite{jiang2024robust} employs joint Gaussians for motion capture and skin Gaussians for appearance details. Although these playback-only techniques achieve high-fidelity rendering in complex scenarios like violin playing, they struggle to generalize to novel motions.

\noindent{\textbf{Animatable Avatar.}} 
Over the past decade, researchers have made significant progress in developing expressive and animatable 3D human avatars.
Early approaches focused on mesh-based methods~\cite{bagautdinov2021driving,xiang2022dressing,xiang2021modeling,halimi2022pattern,casas20144d,shysheya2019textured, deepcap, MonoPerfCap} that utilize motion-controllable template mesh with UV space features to model human geometry and appearance. 
In recent years, research has increasingly shifted towards hybrid representations~\cite{zhu2023trihuman, habermann2023hdhumans} based on implicit models. ARAH~\cite{ARAH:ECCV:2022} utilizes the SDF to model human surfaces. TAVA~\cite{li2022tava} and X-avatar~\cite{shen2023x} learn the skinning weight via Snarf~\cite{chen2021snarf}. DDC~\cite{habermann2021real} employs differentiable rendering to learn the deformation and dynamic texture maps, while deliffas~\cite{kwon2023deliffas} enables fast light field synthesis as a follow-up work. AvatarReX~\cite{zheng2023avatarrex} defines a set of local radiance fields to represent dynamic details of human appearance. Artemis~\cite{luo2022artemis} leverage plenoctree~\cite{yu2021plenoctrees} to represent animal appearances. 
Recent advancements in Gaussian-based representations spark a new era in animatable 3D avatars, leading to the development of innovative techniques~\cite{li2024animatablegaussians, pang2024ash, li2024gaussianbody, qian2023gaussianavatars, chen2024egoavatar, zielonka25dega, rong2024gaussian, zheng2024ohta, zheng2024headgap}.
A bunch of studies
~\cite{lei2024gart,qian20243dgs,hu2023gauhuman} employ linear blend skinning (LBS) to model human motions and reconstruct a 3DGS-based animatable avatar from videos. ASH~\cite{pang2024ash} and GaussianAvatar~\cite{hu2024gaussianavatar} parameterize the 3D character on a 2D UV map while animatableGaussians~\cite{li2024animatablegaussians} relies on the front and back maps to predict Gaussian attributes using 2D CNNs. However, most existing methods focus on human-only scenarios and inherently lack the ability to handle general human-object interactions. In contrast, \methodterm utilizes the Morton-based parameterization and 2D CNNs to learn the attributes of appearance Gaussians, enabling re-performance in complex human-object interactions.

\begin{figure*}[t] 
	\begin{center} 
		\includegraphics[width=\linewidth]{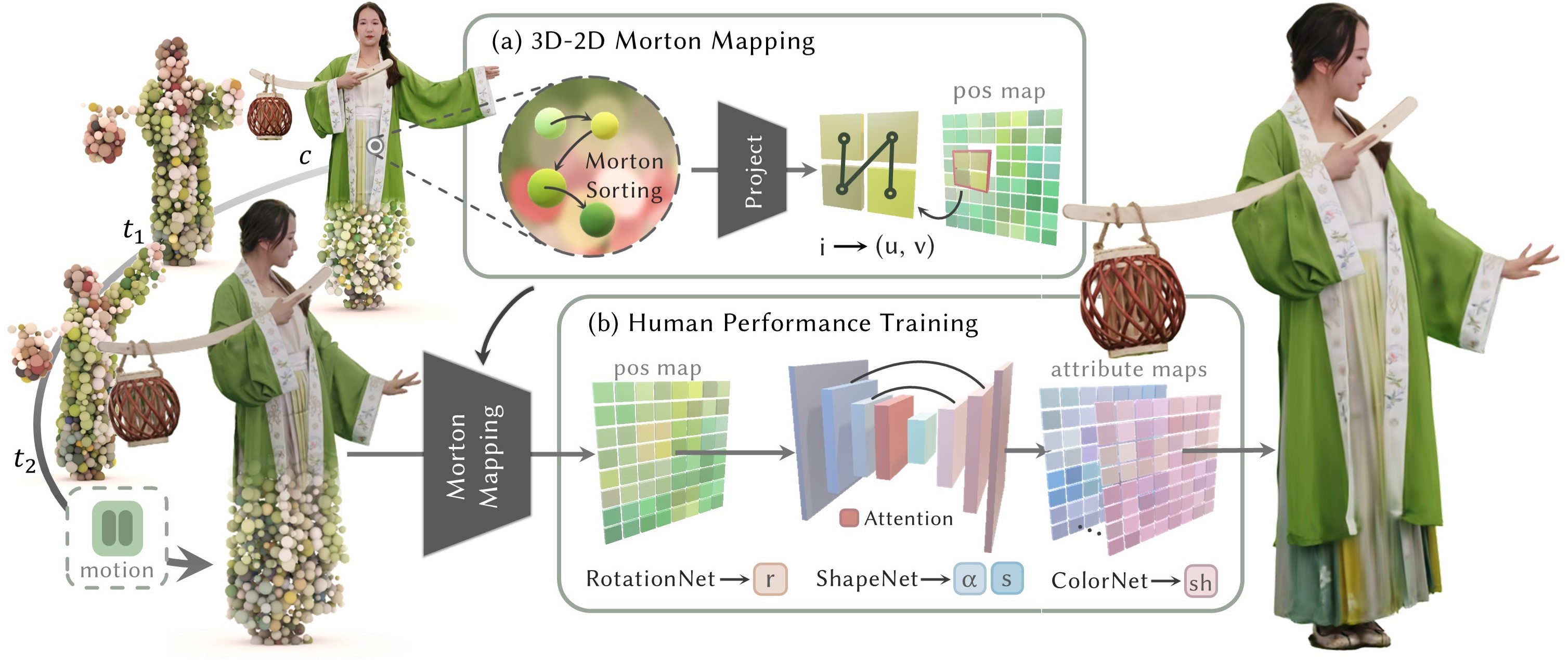} 
	\end{center} 
    \vspace{-20pt}  
	\caption{\noindent{\bf Overview of \methodterm.} Our method disentangles compact motion Gaussians and dense appearance Gaussians, repacking the appearance Gaussians positions into 2D maps for network regression. 
    (a) We use Morton parameterization to project the canonical appearance Gaussians onto UV space, forming a consistent $i \rightarrow (u,v)$ mapping. (b) \methodterm feeds the 2D position map after Morton-based parameterization into three power 2D CNNs with a self-attention layer to regress the corresponding attribute maps. }
	\label{fig:overview} 
	\vspace{-15pt}
\end{figure*}

\section{Method}\label{sec:algorithm} 
Given multi-view human performance videos (up to 81 views), our goal is to generate human-centric volumetric videos that not only deliver high-fidelity playback but also enable vivid re-performance of non-rigid scenes under similar yet unseen motions. The methodology is visually summarized in Fig.~\ref{fig:overview}. Our approach begins by disentangling motion/appearance Gaussians and repacking the attributes of the appearance Gaussians into 2D maps using a Morton-based 2D parameterization. For network training, we adopt the U-Net architecture to learn a generalizable mapping from the position maps to the attribute maps. For re-performance, we adopt a semantic-aware alignment module to associate the motion Gaussians of a new performer with the original appearance Gaussians, enabling seamless transfer and photorealistic rendering. 

\subsection{Motion Tracking and 3D-to-2D Mapping} \label{sec:31}

\noindent{\bf Tracking via Motion Gaussians.}
Inspired by Dynamic 3D Gaussians~\cite{luiten2023dynamic} and SoG~\cite{StollHGST2011}, we utilize a sparse set of 3D Gaussians, denoted as Motion Gaussians, to encapsulate the underlying non-rigid motions $\mathcal{M}$. 
To effectively capture the overall movement of non-rigid scenes with minimal redundancy, we initialize approximately 50,000 Gaussian kernels in the canonical frame $c$ using a uniform random distribution.
To address the issue of insufficient geometric representation in skinny Gaussians, we enforce isotropic and size constraints, resulting in the following initialization loss:
\begin{equation}
E_{\text {init }} = \lambda_{iso} E_{\text {iso }} + \lambda_{size} E_{\text {size }} + E_{\text {color}},
\label{eq:1}
\end{equation}
Please refer to DualGS~\cite{jiang2024robust} for detailed definitions of $E_{\text {iso }}$, $E_{\text {size }}$.
For subsequent frames $t$, we fix spherical harmonics (SH), scaling, and opacity, using differentiable rasterization combined with a local as-rigid-as-possible constraint $E_{\text {arap }}\left(\mathcal{G}^m_{t-1},\mathcal{G}^m_{t}\right)$  to achieve coherent Gaussian movement by optimizing both position $p$ and rotation $q$:
\begin{equation}
\begin{aligned}   
E_{\text {arap }}  = & \sum_i \sum_{k \in \mathcal{N}(i)} w({p_{i,c}^m, p_{k,c}^m}) \| R\left(q^m_{i, t} * {q^m_{i, t-1}}^{-1}\right) \\
& \left(p^m_{k, t-1}-p^m_{i, t-1}\right)-\left(p^m_{k, t}-p^m_{i, t}\right) \|_2^2,
\end{aligned}
\end{equation}
where $m$ denotes the motion Gaussians, $\mathcal{G}^m_{t-1},\mathcal{G}^m_{t}$ are adjacent frame Gaussians, $k$ indicates the nearest 4 neighbors $\mathcal{N}$ of the $i$ Gaussian in the canonical frame. $R(\cdot)$ converts quaternion back into a rotation matrix and $w\left(p_i, p_j\right)=\exp \left(-\left\|p_i-p_j\right\|_2^2 / l^2\right)$ defines the weighting function.

\noindent{\bf Appearance Gaussians.}
Different from sparse Motion Gaussians that provide non-rigid motions, we also utilize dense 3D Gaussians ($\sim$200,000) to serve as the splatting source of final appearance rendering, denoted as appearance Gaussians $\mathcal{T}$.
For initialization of personalized appearance Gaussians, 
recent studies~\cite{li2024animatablegaussians, shao2024degas} typically impose strong regularization that tightly constrains Gaussians to the mesh surface, representing them in the 2D UV space as input for powerful CNNs. However, these constraints heavily compromise rendering quality. In contrast, we train our personalized appearance Gaussians in the canonical frame without additional constraints to preserve high-fidelity details. 

Once trained, given an arbitrary frame $t$ with corresponding motion Gaussians position $p_{k, t}^m$ and rotation $q_{k, t}^m$, we identify the nearest 4 neighbors in the canonical frame $c$ between the dense appearance Gaussians' position $p_{i, c}^\mathcal{T}$ and the sparse motion Gaussians' position $p_{k, c}^m$. We then warp the dense appearance Gaussians $\mathcal{G}^\mathcal{T}_c$ to the target frame denoted as $\mathcal{G}_t^\mathcal{T} = f(\mathcal{G}_c^\mathcal{T}, \mathcal{G}^m_t)$:
\begin{equation}
\begin{aligned}
p_{i, t}^\mathcal{T}=& \sum_{k \in \mathcal{N}\left(p_{i, c}^{\mathcal{T}}\right)} w\left(p_{i, c}^\mathcal{T}, p_{k, c}^m\right)\left(R\left(\Delta q_{k, t}^m\right) p_{i, c}^\mathcal{T}+ \Delta p_{k, t}^m\right), \\
{q}_{i, t}^\mathcal{T} =& \operatorname{slerp}_{k \in \mathcal{N}\left(p_{i, c}^\mathcal{T}\right)} \left( w\left(p_{i, c}^\mathcal{T}, p_{k, c}^m\right), \Delta q_{k, t}^m \right),
\end{aligned}
\end{equation}
$\Delta$ denotes the relative transformation with respect to the canonical frame. We use Slerp to interpolate the rotation.

\noindent{\bf Morton-based 3D-to-2D Mapping.}
The UV atlas of SMPL is inadequate for representing human-object scenarios. To address this, we introduce a new parameterization $\mathcal{P}$ to project the warped appearance Gaussians onto a 2D plane as CNN input, while maintaining strong local pixel-level consistency. 
We first quantize the personalized appearance Gaussians' positions and apply Morton sorting~\cite{morton1966computer} to establish a Morton order, interleaving the binary representations of spatial coordinates to preserve its 3D spatial continuity. 
Each Gaussian $i$ is then assigned a $(u,v)$ coordinate based on this 2D Morton order, forming an $i \rightarrow (u,v)$ mapping, that preserves the adjacency of spatially proximate Gaussians in UV space and remains consistent across all frames.
Using this mapping, we can project the personalized appearance Gaussians warped using motion Gaussians, to form the 2D position map for each frame, where $(u,v)$ coordinates store the corresponding Gaussian positions.

\subsection{Human Performance Training} \label{sec:32}

Unlike DualGS, which overfits appearance Gaussians via gradient descent, we leverage efficient, well-established 2D convolutional architectures to regress the corresponding Gaussian attributes from the 2D position map, enhancing generalization capability. 
Considering the varying attribute distributions, we adopt U-Net~\cite{ronneberger2015u} to learn motion-aware attribute maps, including a Rotation Net for rotation, a Shape Net for opacity and scaling, and a Color Net for the corresponding SH parameters. All attributes 2D maps are then combined with the position map and assembled to form the raw Gaussian representation $\mathcal{G}$. 

However, Morton ordering only preserves local spatial consistency to some extent, leading that two spatially close Gaussian kernels may still be assigned to distant UV coordinates, complicating effective learning for the U-Net. To address this, we deploy a self-attention layer before the final U-Net downsampling step to capture global spatial consistency, complementing the Morton map and further improving the generalization.

\noindent{\bf Training Details.}
We first use background matting~\cite{lin2021real} to extract foreground masks. For motion tracking, we adopt a per-frame training strategy, achieving convergence within 20 seconds for 4,000 iterations across 20 viewpoints. For Morton mapping, We use a $(512,512,3)$ resolution map to store the warped appearance Gaussians positions, filling invalid UV coordinates with zeros.
In the unprojection stage, we extract only valid UV coordinates for attribute maps, assembling the correct Gaussians for subsequent rasterization. 
For each iteration, we randomly sample a frame and a camera viewpoint, generating the corresponding position map to train three specialized U-Nets. This design improves PSNR by 0.25 dB on the DualGS dataset compared to using a single network to regress all attributes.
Before the main training phase, we introduce an additional pre-training stage to warm up the parameters of the three networks. 
In this pre-training stage, we supervise the U-Net by the warped appearance Gaussians as pseudo ground truth, $\mathcal{G}^{\prime}$. This stage optimizes the $\mathcal{L}_2$ loss between the pseudo ground truth and the inferred Gaussian attributes:
\begin{equation}
E_{\text {pre }}=\mathcal{L}_2\left(\mathcal{G},\mathcal{G}^{\prime}\right) ,
\end{equation}
In the main training stage, we assemble the U-Net output attribute maps into Gaussians for rasterization, and supervise using $\mathcal{L}_1$
photometric loss and a D-SSIM term:
\begin{equation}
E_{\text {train }}=\left(1-\lambda_{\text {color }}\right) \mathcal{L}_1+\lambda_{\text {color }} \mathcal{L}_{\text {D-SSIM }},
\end{equation}
We use the Adam optimizer to train the U-Net. In the warm-up stage, we train over 2,000 iterations, with the learning rate linearly decreasing from 1 to 0.1. The main training stage consists of 180,000 iterations. The learning rates for the Rotation Net, Shape Net, and Color Net are initially set to $5 \times 10^{-5}, 5 \times 10^{-4}, 1 \times 10^{-4}$
 , respectively, and linearly decay to one-twentieth of their initial values over the training. The hyperparameters are set to $l = 0.001, \lambda_{iso} = 0.004, \lambda_{size} = 1, \lambda_{color} = 0.2$.

\begin{figure}[tbp] 
	\centering 
	\includegraphics[width=1\linewidth]{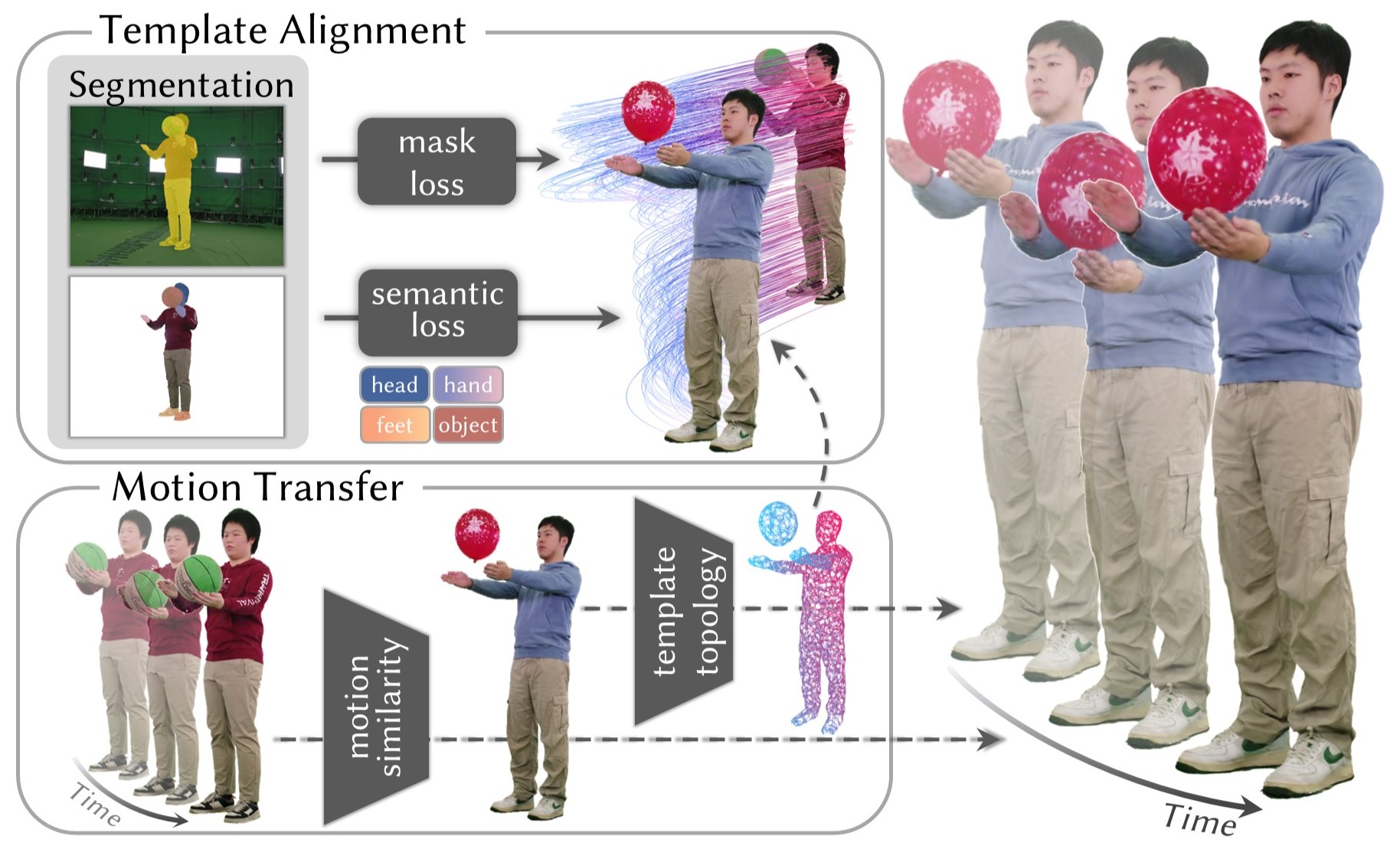} 
	\vspace{-20pt} 
	\caption{The re-performance pipeline consists of two components: template alignment and motion transfer.} 
	\label{fig:animate} 
	\vspace{-15pt} 
\end{figure}

\subsection{Reperformance} \label{sec:33}
Popular parametric model-driven methods struggle with complex human-object interaction scenarios. As shown in Fig.~\ref{fig:animate}, to handle more generalized cases, we propose a semantic-aware motion transfer module based on the motion Gaussian proxy, leveraging the U-Net generalization capability for re-performance. 
For the sequence $s$ to be re-performed, we denote its appearance Gaussians in canonical space as $\mathcal{G}^{s}_{c}$ with positions and rotations $\mathbf{p}_{i,c}^s, \mathbf{q}_{i,c}^s$. 
Similarly, the canonical positions and rotations of the motion Gaussians from the driving sequence $r$ are represented as $\mathbf{p}_{i,c}^r, \mathbf{q}_{i,c}^r$,
with the motion at frame $t$ specified by $\mathbf{p}_{i,t}^r, \mathbf{q}_{i,t }^r$.
Our objective is to recover transformed motions $\mathbf{p}_{i,t}^{s^\prime}, \mathbf{q}_{i,t}^{s^\prime}$ for the appearance Gaussians $\mathcal{G}^{s^\prime}_t$ to be re-performed. As shown in Fig.~\ref{fig:animate}, the motion transfer pipeline includes canonical alignment and motion optimization.

\begin{figure*}[htbp] 
	\begin{center} 
		\includegraphics[width=1.0\linewidth]{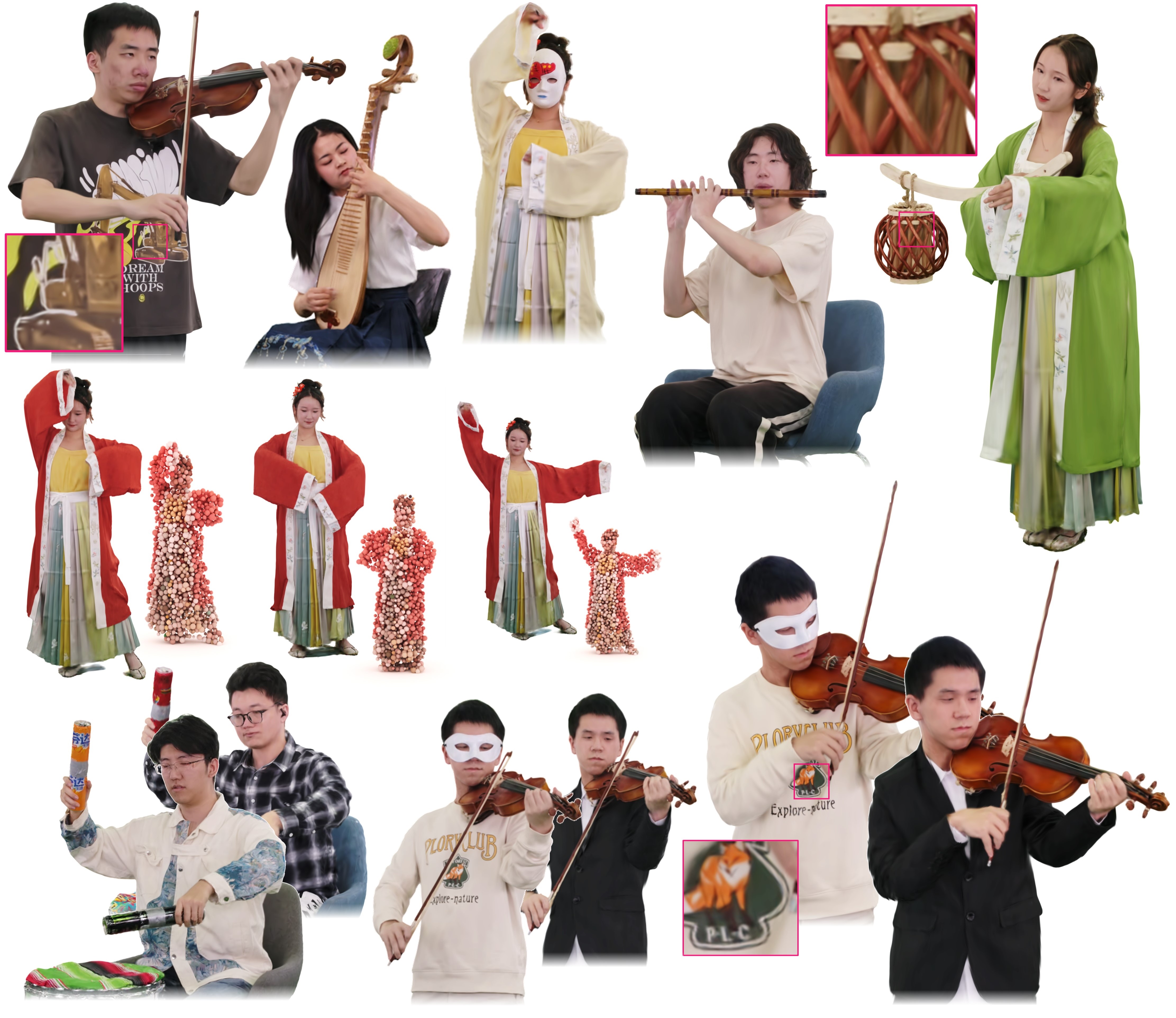} 
	\end{center} 
	\vspace{-20pt}
	\caption{Gallery of our results. \methodterm delivers high-fidelity rendering of human performance in challenging motions and achieves vivid re-performance across various complex human-object scenarios.} 
	\label{fig:gallery}
	\vspace{-20pt}
\end{figure*}

\noindent{\bf Aligment.} 
Pioneering work~\cite{sumner2004deformation} requires users to manually specify corresponding triangle faces between the source mesh and target mesh, indicating that the target triangle with index $i$ should deform like the source triangle with index $j$. However, this approach is infeasible for larger, more complex Gaussian point clouds. 
To address this, we design a semantic-aware supervision method to establish correspondences between $\mathbf{p}_{i,c}^s$ and $\mathbf{p}_{i,c}^r$. Inspired by Language-SAM~\cite{lang-segment-anything}, we use text prompts (e.g., right hand) and leverage GroundingDINO~\cite{liu2023grounding} to generate corresponding bounding boxes, which are then processed by SAM2~\cite{ravi2024sam} to produce semantic masks. Following Semantic Gaussians~\cite{guo2024semantic}, we unproject and fuse the 2D consistent semantic masks onto 3D Gaussians, assigning a semantic label to each Gaussian primitive. 
To establish correspondences, we perform k-means clustering on the Gaussian positions with identical labels for, both, appearance Gaussians to be re-performed and motion Gaussians for driving, respectively. We then optimize the cluster centroids to achieve a coarse alignment of Gaussian position with matching labels. The semantic alignment loss is defined as:
\begin{equation}
E_{\text{sem}} = \sum_{i \in \mathcal{L}} \left\| \mathbf{d}_i^{s'} - \text{sg}[\mathbf{d}_{i}^{r}] \right\|_2^2,
\end{equation}
where $\mathcal{L}$ is the set of semantic labels, including head, hands, feet, and objects(e.g., balloon). $\mathbf{d}_i^{s'}$ and $\mathbf{d}_{i}^{r}$ are the centroids of the optimized source Gaussians and target for label $i$. $sg$ stands for the stop-gradient operator.
The overall alignment loss is defined as:
\begin{equation} 
E_{\text{align}} = \left\| \hat{\mathcal{M}} - \mathcal{M}_r \right\|_1 + \lambda_{\text{sem}} E_{\text{sem}} + \lambda_{\text{1}} E_{\text{arap}}\left(\mathcal{G}^{s}_{c},\mathcal{G}^{s^\prime}_{c}\right), 
\label{eq:align}
\end{equation}

where $\hat{\mathcal{M}}$ is the alpha channel obtained from rendering the Gaussian $\mathcal{G}^{s^\prime}_{c}$, and $\mathcal{M}_r$ is the ground truth mask.
The mask loss and semantic loss ensure silhouette alignment, while the arap loss preserves the topology of the appearance Gaussians to be re-performed.

\noindent{\bf Motion Transfer.} 
To transfer the motion Gaussians deformations while preserving the topology of the appearance Gaussians to be re-performed, we optimize the position and rotation of the appearance Gaussians $\mathcal{G}^{s^\prime}_t$  under the novel motion of $\mathcal{G}^r_t$. The optimization objective is defined as:
\begin{equation} 
E_{\text {re}} =  \mathcal{L}_2\left(\mathcal{G}^{s^\prime}_t,  f(\mathcal{G}^{s^\prime}_{c},\mathcal{G}^r_t) \right)+ \lambda_{2}
E_{\text {arap }}\left(\mathcal{G}^{s}_{c}, \mathcal{G}^{s^\prime}_{t}\right).
\label{eq:transfer}
\end{equation}
where the first term minimizes the difference between the source motion is similar to the driven motion, and the second term enforces the as-rigid-as-possible constraint to maintain the original topology of the appearance Gaussians. 
The optimized $\mathbf{p}_{i,t}^{s^\prime}$ forms a new position map, retaining the same Morton order, which is then fed into the trained U-Net to recover the final re-performed Gaussians.
Note that the hyperparameters are $l = 0.001, \lambda_{sem} = 0.001, \lambda_{1} = 0.01, \lambda_{2} =  0.2$.

\newpage

\begin{figure*}[htbp] 
	\begin{center} 
		\includegraphics[width=0.99\linewidth]{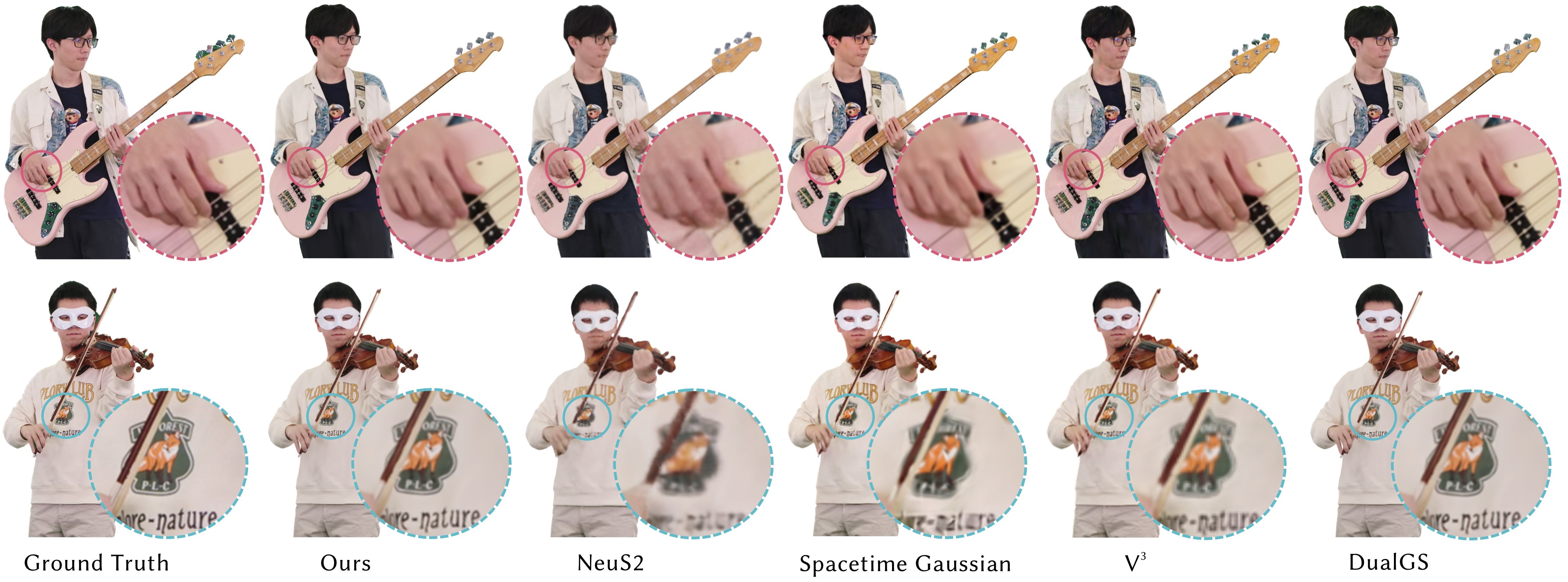} 
	\end{center} 
	\vspace{-16pt}
	\caption{Qualitative comparison with SOTA playback-only methods on novel view synthesis on DualGS dataset.}
	\label{fig:fig_comp_1}
\end{figure*}

\begin{figure*}[htbp] 
	\begin{center} 
		\includegraphics[width=0.99\linewidth]{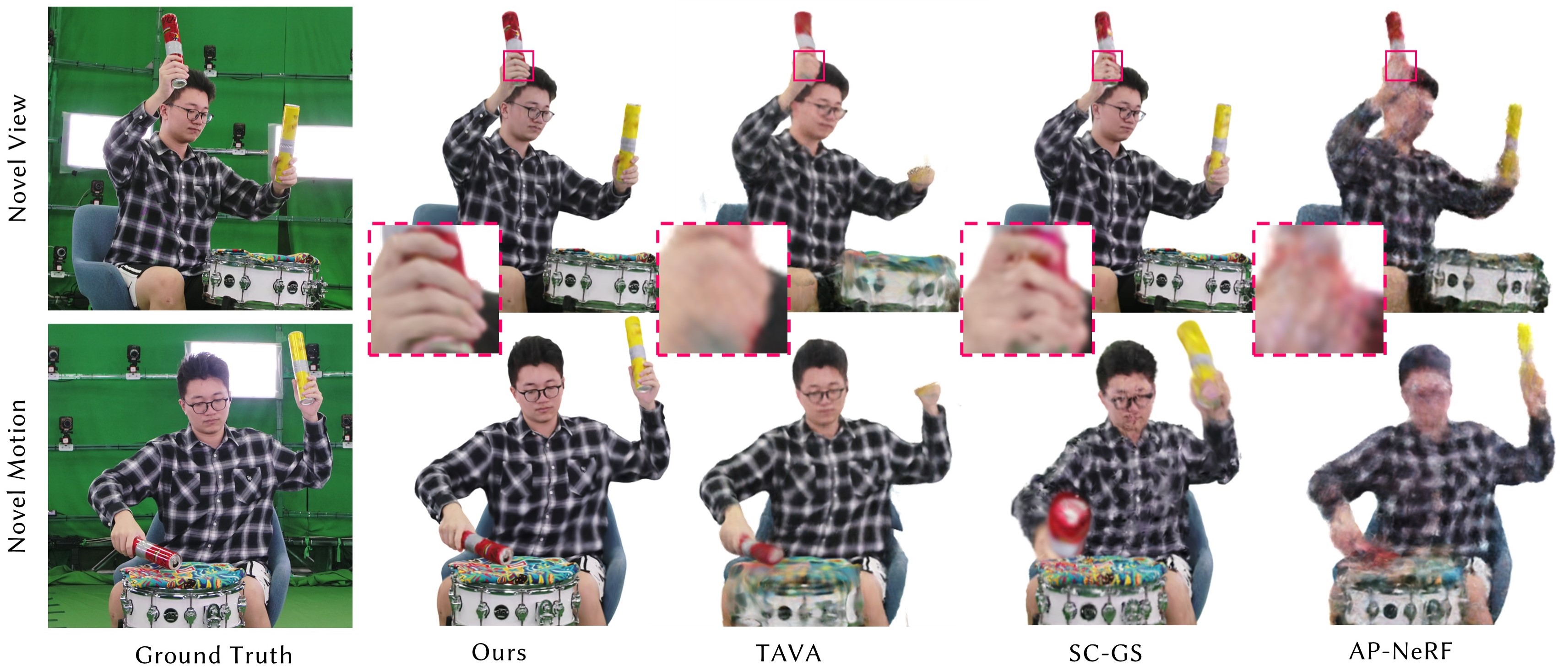} 
	\end{center} 
	\vspace{-16pt}
	\caption{Qualitative comparison with AP-NeRF~\cite{uzolas2024template}, TAVA~\cite{li2022tava} and SC-GS~\cite{huang2024sc} on both training motion and novel motion synthesis.}
        \vspace{-16pt}
	\label{fig:fig_comp_2}
\end{figure*}

\section{Experimental Results} 
\subsection{Comparison} 
\myparagraph{Qualitative Evaluation.} We compare our \methodterm against SOTA dynamic rendering methods, including NeuS2~\cite{wang2023neus2}, Spacetime Gaussian~\cite{li2023spacetime}, $V^3$~\cite{wang2024v} and DualGS~\cite{jiang2024robust}, using the DualGS dataset. As shown in Fig.~\ref{fig:fig_comp_1}, NeuS2 produces blurry results, while Spacetime Gaussian and $V^3$, which sacrifices high-order SH for faster training, suffer from oversmoothing and a lack of rich details.
Although DualGS achieves high-quality rendering, its per-frame optimization approach is computationally intensive. 
Yet, such methods are limited to playback and inherently lack the ability to generalize to new motions.

To validate generalization capability, we note that existing methods cannot effectively handle re-performance in general human-object interaction scenarios. Thus, we select control-point-driven and template-free modeling approaches, training them in a multi-view setting to evaluate their rendering quality on novel views and motions. As depicted in Fig.~\ref{fig:fig_comp_2}, Articulated-Point-NeRF~\cite{uzolas2024template}, which relies heavily on TiNeuVox~\cite{TiNeuVox}, exhibits significant artifacts when training on complex, long sequences.
Similarly, TAVA~\cite{li2022tava} faces difficulties with challenging motions, failing to capture fine appearance details, while SC-GS~\cite{huang2024sc} achieves reasonable rendering quality for novel views but suffers severe degradation in appearance quality with novel poses.
In contrast, \methodterm leverages Morton-based parameterization and a U-Net with self-attention to ensure 2D consistency, enabling high-fidelity rendering for, both, novel views and novel motions.
As illustrated in Fig.~\ref{fig:gallery}, our method effectively addresses the dual challenges of playback and re-performance across various complex human-object scenarios.

\myparagraph{Quantitative Comparison.}
We select three 500-frame sequences from the DualGS dataset and evaluate performance across 8 test views using metrics such as PSNR, SSIM, LPIPS, and training time.
As shown in Tab.~\ref{table:Qualitativecomparison}, \methodterm ranks second, slightly behind per-frame optimized DualGS, with significantly reduced training time thanks to 2D CNNs' generalization ability.

To assess generalization, we use a 3,000-frame sequence, training on frames 1 to 2,500 and testing on frames 2,500 to 3,000. As illustrated in Tab.~\ref{table:Qualitativecomparison2}, \methodterm outperforms all other methods across multiple metrics, demonstrating superior rendering quality and strong generalization capability.

\begin{table}[t]
	\begin{center}
		\centering
		\vspace{20pt}

		\resizebox{0.47\textwidth}{!}{
			\begin{tabular}{l|cccc}
				\hline
				Method   &  PSNR $\uparrow$ & SSIM $\uparrow$ & LPIPS $\downarrow$ & Training Time(Min / frame) $\downarrow$ \\
				\hline
				NeuS2~\cite{wang2023neus2}\qquad\qquad & 29.589 & 0.967 & 0.0561 & 3.23  \\
			    Spacetime Gaussian~\cite{li2023spacetime}     & 31.691 & 0.981 & 0.0289 & 2.24  \\
                DualGS~\cite{jiang2024robust} & \bestCellColor{35.506} & \bestCellColor{0.990} & \secondBestCellColor{0.0193} & 12.22  \\
                $V^3$~\cite{wang2024v} & 33.821 & 0.984 & \bestCellColor{0.0167} & \secondBestCellColor{\textbf{1.86}}  \\
    		\hline
                Ours & \secondBestCellColor{\textbf{34.571}} & \secondBestCellColor{\textbf{0.986}} & \textbf{0.0229} & \bestCellColor{\textbf{1.68}} \\
				\hline
			\end{tabular}
            }
            \caption{
            Quantitative comparison of novel view rendering. Green and yellow cell colors indicate the best and the second-best results. %
            }
		\label{table:Qualitativecomparison}
        \vspace{-20pt}
	\end{center}
\end{table}

\begin{table}[t]
    \centering

    \resizebox{\columnwidth}{!}{ %
    \begin{tabular}{c|c c c|c c c}
        \hline
        \multirow{2}{*}{\textbf{Methods}} & \multicolumn{3}{c|}{\textbf{Novel View}} & \multicolumn{3}{c}{\textbf{Novel Motion}} \\
        & PSNR $\uparrow$ & SSIM $\uparrow$  &  LPIPS $\downarrow$ & PSNR $\uparrow$ & SSIM $\uparrow$  &  LPIPS $\downarrow$ \\
        \hline
        AP-NeRF ~\cite{uzolas2024template} & 28.26 & 0.939 & 0.0768 & \secondBestCellColor{26.85} & \secondBestCellColor{0.944} & \secondBestCellColor{0.0519} \\
        TAVA~\cite{li2022tava}  & 21.57 & 0.906 &  0.1292& 21.34 & 0.905 & 0.1301 \\
        SC-GS~\cite{huang2024sc}  & \secondBestCellColor{30.63} & \secondBestCellColor{0.971} & \secondBestCellColor{0.0319} & 19.18 & 0.907 & 0.0821 \\
        \hline

        Ours  & \bestCellColor{\textbf{32.09}} & \bestCellColor{\textbf{0.979}} & \bestCellColor{\textbf{0.0310}} & \bestCellColor{\textbf{30.06}} & \bestCellColor{\textbf{0.976}} & \bestCellColor{\textbf{0.0277}} \\
        \hline
    \end{tabular}
    }
    \caption{
    Quantitative comparison with template-free modeling approaches in both novel views and motions. 
    }
    \label{table:Qualitativecomparison2}
\end{table}

\subsection{Ablation Study} \label{sec:abla} 

\begin{figure}[t] 
	\begin{center} 
		\includegraphics[width=1\linewidth]{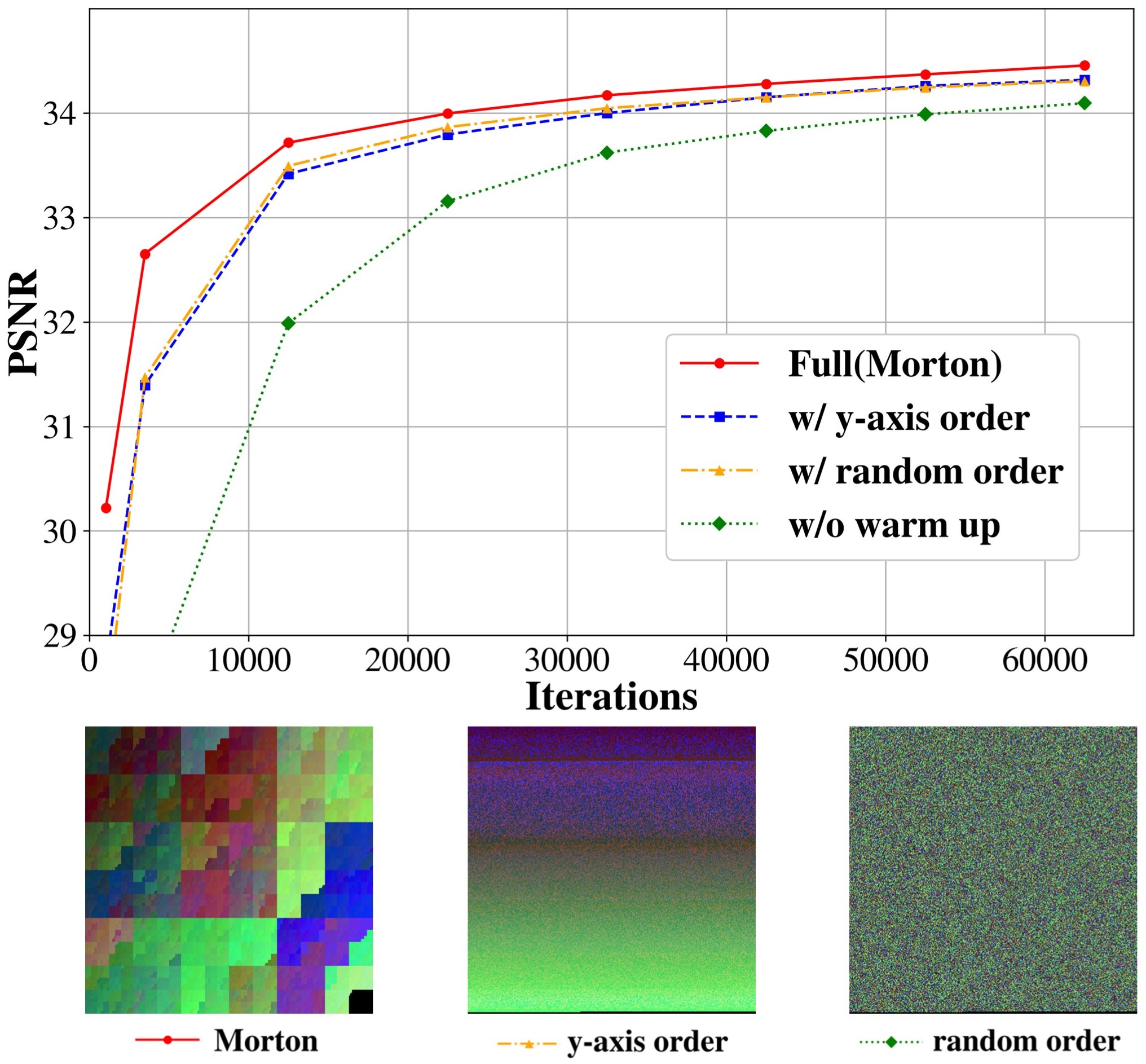}
	\end{center} 
	\vspace{-14pt}
	\caption{Qualitative ablation study on training strategy. Our full pipeline achieves the fastest convergence and highest quality.} 
    \vspace{-15pt}
	\label{fig:training_strategy}
\end{figure}

\myparagraph{Training Strategy.} 
We conduct a qualitative ablation study on 500 frames, comparing different 3D-to-2D parameterization methods: Morton, y-axis order, and random mapping, to evaluate their effectiveness. As shown in Fig.~\ref{fig:training_strategy}, Morton parameterization outperforms both y-axis and random mappings in convergence speed and rendering quality. The bottom images show position maps normalized to the RGB color space, demonstrating Morton mapping that preserves local consistency in UV space. Additionally, skipping the warm-up stage leads to slower convergence during training.

\begin{figure}[t] 
        \vspace{-10pt}
	\begin{center} 
		\includegraphics[width=1\linewidth]{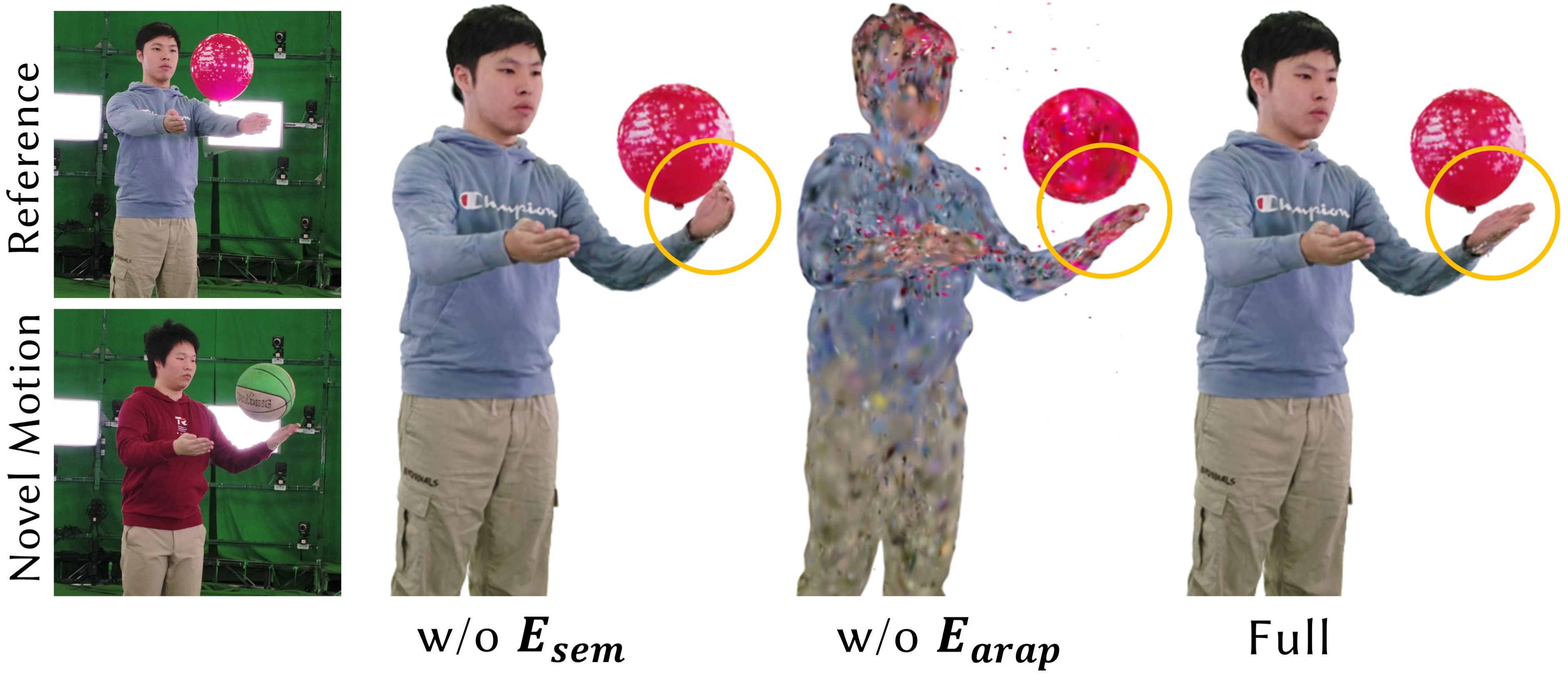} 
	\end{center} 
    \vspace{-14pt}
	\caption{Qualitative ablation study on alignment component.} 
	\label{fig:alignment_ablation}
	\vspace{-10pt}
\end{figure}

\myparagraph{Aligment.} We assess the impact of each term in Eq.~\ref{eq:align}. As depicted in Fig.~\ref{fig:alignment_ablation}, excluding $E_{\text{sem}}$ leads to unnatural artifacts, while omitting $E_{\text{arap}}$ results in severe noise. In contrast, our complete method successfully aligns the source with the target, while maintaining topological consistency.

\begin{figure}[t] 
	\vspace{10pt}

	\begin{center} 
		\includegraphics[width=1\linewidth]{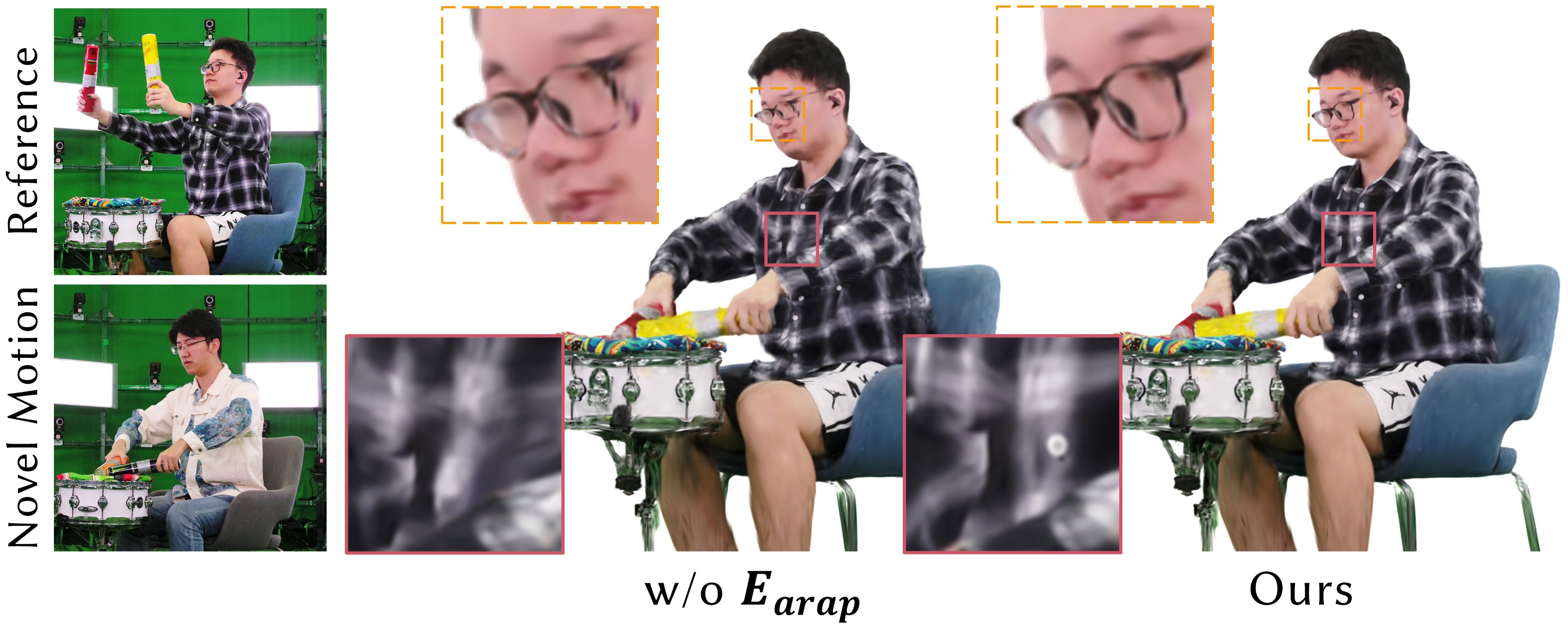}
	\end{center} 
    \vspace{-14pt}
	\caption{Qualitative ablation study on motion transfer.} 
	\label{fig:motiontransfer_ablation}
	\vspace{-15pt}
\end{figure}

\myparagraph{Motion Transfer.} We analyze the influence of $E_{\text{arap}}$ in Eq.~\ref{eq:transfer}, examining the trade-off between motion similarity and topological consistency. As shown in Fig.~\ref{fig:motiontransfer_ablation}, excluding $E_{\text{arap}}$ causes unrealistic deformations and artifacts in the appearance Gaussians, even after U-Net processing. Including the $E_{\text{arap}}$ term significantly alleviates these issues, leading to more realistic rendering for novel motions.


\section{Limitations}
Despite \methodterm achieving high-fidelity rendering for novel views and motions, it has several limitations.
First, although our motion Gaussians effectively capture dynamics in non-rigid scenes, the process remains time-consuming (~20 seconds per frame), limiting real-time applications. Accelerating this step is a promising direction for future research.
Second, \methodterm struggles to generalize across arbitrary motions. The quality of re-performance relies heavily on having similar interaction types, collecting large-scale datasets could enhance generalization.
Finally, we utilized only one self-attention layer due to GPU memory constraints. Using additional layers could further enhance generalization.

\section {Conclusion} 
We have introduced a Gaussian-based approach that not only delivers high-fidelity playback but also enables vivid re-performance of general non-rigid scenes.
Our approach disentangles general non-rigid scenes into compact motion Gaussians and dense appearance Gaussians, and employs a Morton-based parameterization to repack the appearance Gaussians into 2D position maps. 
We utilize the U-Net architecture to regress the corresponding attribute maps of the position maps, which are assembled into appearance Gaussians for photorealistic rendering. 
We also introduce a semantic-aware alignment module that transfers motion Gaussians from a new performer to the original scene for re-performance.
Comprehensive experiments demonstrate that \methodterm outperforms state-of-the-art methods in rendering quality and generalization capability.
With the unique and faithful playback and re-performance ability, we believe our approach can serve as a solid step towards immersive human-centric volumetric videos.

\noindent{\bf Acknowledgements.} This work was supported by the National Key R$\&$D Program of China (2022YFF0902301), Shanghai Local College Capacity Building Program (22010502800), and Shanghai Frontiers Science Center of Human-centered AI (ShangHAI).

{
    \small
    \bibliographystyle{ieeenat_fullname}
    \bibliography{reference}

\begin{thebibliography}{90}
\providecommand{\natexlab}[1]{#1}
\providecommand{\url}[1]{\texttt{#1}}
\expandafter\ifx\csname urlstyle\endcsname\relax
  \providecommand{\doi}[1]{doi: #1}\else
  \providecommand{\doi}{doi: \begingroup \urlstyle{rm}\Url}\fi

\bibitem[Bagautdinov et~al.(2021)Bagautdinov, Wu, Simon, Prada, Shiratori, Wei,
  Xu, Sheikh, and Saragih]{bagautdinov2021driving}
Timur Bagautdinov, Chenglei Wu, Tomas Simon, Fabian Prada, Takaaki Shiratori,
  Shih-En Wei, Weipeng Xu, Yaser Sheikh, and Jason Saragih.
\newblock Driving-signal aware full-body avatars.
\newblock \emph{ACM Transactions on Graphics (TOG)}, 40\penalty0 (4):\penalty0
  1--17, 2021.

\bibitem[Cao and Johnson(2023)]{cao2023hexplane}
Ang Cao and Justin Johnson.
\newblock Hexplane: A fast representation for dynamic scenes.
\newblock In \emph{Proceedings of the IEEE/CVF Conference on Computer Vision
  and Pattern Recognition}, pages 130--141, 2023.

\bibitem[Casas et~al.(2014)Casas, Volino, Collomosse, and Hilton]{casas20144d}
Dan Casas, Marco Volino, John Collomosse, and Adrian Hilton.
\newblock 4d video textures for interactive character appearance.
\newblock In \emph{Computer Graphics Forum}, pages 371--380. Wiley Online
  Library, 2014.

\bibitem[Chen et~al.(2022)Chen, Xu, Geiger, Yu, and Su]{chen2022tensorf}
Anpei Chen, Zexiang Xu, Andreas Geiger, Jingyi Yu, and Hao Su.
\newblock Tensorf: Tensorial radiance fields.
\newblock In \emph{European Conference on Computer Vision}, pages 333--350.
  Springer, 2022.

\bibitem[Chen et~al.(2024)Chen, Wang, Zhang, Pandey, Beeler, Habermann, and
  Theobalt]{chen2024egoavatar}
Jianchun Chen, Jian Wang, Yinda Zhang, Rohit Pandey, Thabo Beeler, Marc
  Habermann, and Christian Theobalt.
\newblock Egoavatar: Egocentric view-driven and photorealistic full-body
  avatars.
\newblock In \emph{SIGGRAPH Asia 2024 Conference Papers}, pages 1--11, 2024.

\bibitem[Chen et~al.(2021)Chen, Zheng, Black, Hilliges, and
  Geiger]{chen2021snarf}
Xu Chen, Yufeng Zheng, Michael~J Black, Otmar Hilliges, and Andreas Geiger.
\newblock Snarf: Differentiable forward skinning for animating non-rigid neural
  implicit shapes.
\newblock In \emph{Proceedings of the IEEE/CVF International Conference on
  Computer Vision}, pages 11594--11604, 2021.

\bibitem[Collet et~al.(2015)Collet, Chuang, Sweeney, Gillett, Evseev,
  Calabrese, Hoppe, Kirk, and Sullivan]{collet2015high}
Alvaro Collet, Ming Chuang, Pat Sweeney, Don Gillett, Dennis Evseev, David
  Calabrese, Hugues Hoppe, Adam Kirk, and Steve Sullivan.
\newblock High-quality streamable free-viewpoint video.
\newblock \emph{ACM Transactions on Graphics (TOG)}, 34\penalty0 (4):\penalty0
  69, 2015.

\bibitem[Dou et~al.(2016)Dou, Khamis, Degtyarev, Davidson, Fanello, Kowdle,
  Escolano, Rhemann, Kim, Taylor, Kohli, Tankovich, and Izadi]{dou2016fusion4d}
Mingsong Dou, Sameh Khamis, Yury Degtyarev, Philip Davidson, Sean~Ryan Fanello,
  Adarsh Kowdle, Sergio~Orts Escolano, Christoph Rhemann, David Kim, Jonathan
  Taylor, Pushmeet Kohli, Vladimir Tankovich, and Shahram Izadi.
\newblock Fusion4d: real-time performance capture of challenging scenes.
\newblock \emph{ACM Trans. Graph.}, 35\penalty0 (4), 2016.

\bibitem[Dou et~al.(2017)Dou, Davidson, Fanello, Khamis, Kowdle, Rhemann,
  Tankovich, and Izadi]{motion2fusion}
Mingsong Dou, Philip Davidson, Sean~Ryan Fanello, Sameh Khamis, Adarsh Kowdle,
  Christoph Rhemann, Vladimir Tankovich, and Shahram Izadi.
\newblock Motion2fusion: Real-time volumetric performance capture.
\newblock \emph{ACM Trans. Graph.}, 36\penalty0 (6):\penalty0 246:1--246:16,
  2017.

\bibitem[Duan et~al.(2024)Duan, Wei, Dai, He, Chen, and Chen]{duan20244d}
Yuanxing Duan, Fangyin Wei, Qiyu Dai, Yuhang He, Wenzheng Chen, and Baoquan
  Chen.
\newblock 4d-rotor gaussian splatting: towards efficient novel view synthesis
  for dynamic scenes.
\newblock In \emph{ACM SIGGRAPH 2024 Conference Papers}, pages 1--11, 2024.

\bibitem[Fang et~al.(2022)Fang, Yi, Wang, Xie, Zhang, Liu, Nie\ss{}ner, and
  Tian]{TiNeuVox}
Jiemin Fang, Taoran Yi, Xinggang Wang, Lingxi Xie, Xiaopeng Zhang, Wenyu Liu,
  Matthias Nie\ss{}ner, and Qi Tian.
\newblock Fast dynamic radiance fields with time-aware neural voxels.
\newblock In \emph{SIGGRAPH Asia 2022 Conference Papers}, 2022.

\bibitem[Fridovich-Keil et~al.(2023)Fridovich-Keil, Meanti, Warburg, Recht, and
  Kanazawa]{fridovich2023k}
Sara Fridovich-Keil, Giacomo Meanti, Frederik~Rahb{\ae}k Warburg, Benjamin
  Recht, and Angjoo Kanazawa.
\newblock K-planes: Explicit radiance fields in space, time, and appearance.
\newblock In \emph{Proceedings of the IEEE/CVF Conference on Computer Vision
  and Pattern Recognition}, pages 12479--12488, 2023.

\bibitem[Gao et~al.(2024)Gao, Xu, Cao, Mildenhall, Ma, Chen, Tang, and
  Neumann]{gao2024gaussianflow}
Quankai Gao, Qiangeng Xu, Zhe Cao, Ben Mildenhall, Wenchao Ma, Le Chen, Danhang
  Tang, and Ulrich Neumann.
\newblock Gaussianflow: Splatting gaussian dynamics for 4d content creation.
\newblock \emph{arXiv preprint arXiv:2403.12365}, 2024.

\bibitem[Guo et~al.(2024{\natexlab{a}})Guo, Ma, Fan, Liu, and
  Li]{guo2024semantic}
Jun Guo, Xiaojian Ma, Yue Fan, Huaping Liu, and Qing Li.
\newblock Semantic gaussians: Open-vocabulary scene understanding with 3d
  gaussian splatting.
\newblock \emph{arXiv preprint arXiv:2403.15624}, 2024{\natexlab{a}}.

\bibitem[Guo et~al.(2024{\natexlab{b}})Guo, Zhou, Li, Wang, and
  Li]{guo2024motion}
Zhiyang Guo, Wengang Zhou, Li Li, Min Wang, and Houqiang Li.
\newblock Motion-aware 3d gaussian splatting for efficient dynamic scene
  reconstruction.
\newblock \emph{IEEE Transactions on Circuits and Systems for Video
  Technology}, 2024{\natexlab{b}}.

\bibitem[Habermann et~al.(2020)Habermann, Xu, Zollhoefer, Pons-Moll, and
  Theobalt]{deepcap}
Marc Habermann, Weipeng Xu, Michael Zollhoefer, Gerard Pons-Moll, and Christian
  Theobalt.
\newblock Deepcap: Monocular human performance capture using weak supervision.
\newblock In \emph{{IEEE} Conference on Computer Vision and Pattern Recognition
  (CVPR)}. {IEEE}, 2020.

\bibitem[Habermann et~al.(2021)Habermann, Liu, Xu, Zollhoefer, Pons-Moll, and
  Theobalt]{habermann2021real}
Marc Habermann, Lingjie Liu, Weipeng Xu, Michael Zollhoefer, Gerard Pons-Moll,
  and Christian Theobalt.
\newblock Real-time deep dynamic characters.
\newblock \emph{ACM Transactions on Graphics (ToG)}, 40\penalty0 (4):\penalty0
  1--16, 2021.

\bibitem[Habermann et~al.(2023)Habermann, Liu, Xu, Pons-Moll, Zollhoefer, and
  Theobalt]{habermann2023hdhumans}
Marc Habermann, Lingjie Liu, Weipeng Xu, Gerard Pons-Moll, Michael Zollhoefer,
  and Christian Theobalt.
\newblock Hdhumans: A hybrid approach for high-fidelity digital humans.
\newblock \emph{Proceedings of the ACM on Computer Graphics and Interactive
  Techniques}, 6\penalty0 (3):\penalty0 1--23, 2023.

\bibitem[Halimi et~al.(2022)Halimi, Stuyck, Xiang, Bagautdinov, Wen, Kimmel,
  Shiratori, Wu, Sheikh, and Prada]{halimi2022pattern}
Oshri Halimi, Tuur Stuyck, Donglai Xiang, Timur~M Bagautdinov, He Wen, Ron
  Kimmel, Takaaki Shiratori, Chenglei Wu, Yaser Sheikh, and Fabian Prada.
\newblock Pattern-based cloth registration and sparse-view animation.
\newblock \emph{ACM Trans. Graph.}, 41\penalty0 (6):\penalty0 196--1, 2022.

\bibitem[Hu et~al.(2024{\natexlab{a}})Hu, Zhang, Zhang, Zhou, Liu, Zhang, and
  Nie]{hu2024gaussianavatar}
Liangxiao Hu, Hongwen Zhang, Yuxiang Zhang, Boyao Zhou, Boning Liu, Shengping
  Zhang, and Liqiang Nie.
\newblock Gaussianavatar: Towards realistic human avatar modeling from a single
  video via animatable 3d gaussians.
\newblock In \emph{Proceedings of the IEEE/CVF Conference on Computer Vision
  and Pattern Recognition}, pages 634--644, 2024{\natexlab{a}}.

\bibitem[Hu et~al.(2024{\natexlab{b}})Hu, Hu, and Liu]{hu2023gauhuman}
Shoukang Hu, Tao Hu, and Ziwei Liu.
\newblock Gauhuman: Articulated gaussian splatting from monocular human videos.
\newblock In \emph{Proceedings of the IEEE/CVF Conference on Computer Vision
  and Pattern Recognition}, pages 20418--20431, 2024{\natexlab{b}}.

\bibitem[Huang et~al.(2024)Huang, Sun, Yang, Lyu, Cao, and Qi]{huang2024sc}
Yi-Hua Huang, Yang-Tian Sun, Ziyi Yang, Xiaoyang Lyu, Yan-Pei Cao, and Xiaojuan
  Qi.
\newblock Sc-gs: Sparse-controlled gaussian splatting for editable dynamic
  scenes.
\newblock In \emph{Proceedings of the IEEE/CVF Conference on Computer Vision
  and Pattern Recognition}, pages 4220--4230, 2024.

\bibitem[I\c{s}{\i}k et~al.(2023)I\c{s}{\i}k, Rünz, Georgopoulos, Khakhulin,
  Starck, Agapito, and Nießner]{isik2023humanrf}
Mustafa I\c{s}{\i}k, Martin Rünz, Markos Georgopoulos, Taras Khakhulin,
  Jonathan Starck, Lourdes Agapito, and Matthias Nießner.
\newblock Humanrf: High-fidelity neural radiance fields for humans in motion.
\newblock \emph{ACM Transactions on Graphics (TOG)}, 42\penalty0 (4):\penalty0
  1--12, 2023.

\bibitem[Jiang et~al.(2023{\natexlab{a}})Jiang, Chen, Song, and
  Hilliges]{jiang2023instantavatar}
Tianjian Jiang, Xu Chen, Jie Song, and Otmar Hilliges.
\newblock Instantavatar: Learning avatars from monocular video in 60 seconds.
\newblock In \emph{Proceedings of the IEEE/CVF Conference on Computer Vision
  and Pattern Recognition}, pages 16922--16932, 2023{\natexlab{a}}.

\bibitem[Jiang et~al.(2022)Jiang, Jiang, Sun, Su, Guo, Wu, Yu, and
  Xu]{jiang2022neuralhofusion}
Yuheng Jiang, Suyi Jiang, Guoxing Sun, Zhuo Su, Kaiwen Guo, Minye Wu, Jingyi
  Yu, and Lan Xu.
\newblock Neuralhofusion: Neural volumetric rendering under human-object
  interactions.
\newblock In \emph{Proceedings of the IEEE/CVF Conference on Computer Vision
  and Pattern Recognition}, pages 6155--6165, 2022.

\bibitem[Jiang et~al.(2023{\natexlab{b}})Jiang, Yao, Su, Shen, Luo, and
  Xu]{jiang2023instant}
Yuheng Jiang, Kaixin Yao, Zhuo Su, Zhehao Shen, Haimin Luo, and Lan Xu.
\newblock Instant-nvr: Instant neural volumetric rendering for human-object
  interactions from monocular rgbd stream.
\newblock In \emph{Proceedings of the IEEE/CVF Conference on Computer Vision
  and Pattern Recognition}, pages 595--605, 2023{\natexlab{b}}.

\bibitem[Jiang et~al.(2024{\natexlab{a}})Jiang, Shen, Hong, Guo, Wu, Zhang, Yu,
  and Xu]{jiang2024robust}
Yuheng Jiang, Zhehao Shen, Yu Hong, Chengcheng Guo, Yize Wu, Yingliang Zhang,
  Jingyi Yu, and Lan Xu.
\newblock Robust dual gaussian splatting for immersive human-centric volumetric
  videos.
\newblock \emph{ACM Transactions on Graphics (TOG)}, 43\penalty0 (6):\penalty0
  1--15, 2024{\natexlab{a}}.

\bibitem[Jiang et~al.(2024{\natexlab{b}})Jiang, Shen, Wang, Su, Hong, Zhang,
  Yu, and Xu]{jiang2024hifi4g}
Yuheng Jiang, Zhehao Shen, Penghao Wang, Zhuo Su, Yu Hong, Yingliang Zhang,
  Jingyi Yu, and Lan Xu.
\newblock Hifi4g: High-fidelity human performance rendering via compact
  gaussian splatting.
\newblock In \emph{Proceedings of the IEEE/CVF Conference on Computer Vision
  and Pattern Recognition}, pages 19734--19745, 2024{\natexlab{b}}.

\bibitem[Joo et~al.(2018)Joo, Simon, and Sheikh]{TotalCapture}
Hanbyul Joo, Tomas Simon, and Yaser Sheikh.
\newblock Total capture: A 3d deformation model for tracking faces, hands, and
  bodies.
\newblock In \emph{The IEEE Conference on Computer Vision and Pattern
  Recognition (CVPR)}, 2018.

\bibitem[Kerbl et~al.(2023)Kerbl, Kopanas, Leimk{\"u}hler, and
  Drettakis]{kerbl20233d}
Bernhard Kerbl, Georgios Kopanas, Thomas Leimk{\"u}hler, and George Drettakis.
\newblock 3d gaussian splatting for real-time radiance field rendering.
\newblock \emph{ACM Transactions on Graphics (ToG)}, 42\penalty0 (4):\penalty0
  1--14, 2023.

\bibitem[Kwon et~al.(2023)Kwon, Liu, Fuchs, Habermann, and
  Theobalt]{kwon2023deliffas}
Youngjoong Kwon, Lingjie Liu, Henry Fuchs, Marc Habermann, and Christian
  Theobalt.
\newblock Deliffas: Deformable light fields for fast avatar synthesis.
\newblock \emph{Advances in Neural Information Processing Systems}, 2023.

\bibitem[Labe et~al.(2024)Labe, Issachar, Lang, and Benaim]{labe2024dgd}
Isaac Labe, Noam Issachar, Itai Lang, and Sagie Benaim.
\newblock Dgd: Dynamic 3d gaussians distillation.
\newblock In \emph{European Conference on Computer Vision}, pages 361--378.
  Springer, 2024.

\bibitem[Lei et~al.(2024)Lei, Wang, Pavlakos, Liu, and Daniilidis]{lei2024gart}
Jiahui Lei, Yufu Wang, Georgios Pavlakos, Lingjie Liu, and Kostas Daniilidis.
\newblock Gart: Gaussian articulated template models.
\newblock In \emph{Proceedings of the IEEE/CVF Conference on Computer Vision
  and Pattern Recognition}, pages 19876--19887, 2024.

\bibitem[Li et~al.(2008)Li, Sumner, and Pauly]{li2008global}
Hao Li, Robert~W Sumner, and Mark Pauly.
\newblock Global correspondence optimization for non-rigid registration of
  depth scans.
\newblock In \emph{Computer graphics forum}, pages 1421--1430. Wiley Online
  Library, 2008.

\bibitem[Li et~al.(2024{\natexlab{a}})Li, Yao, Xie, Chen, and
  Jiang]{li2024gaussianbody}
Mengtian Li, Shengxiang Yao, Zhifeng Xie, Keyu Chen, and Yu-Gang Jiang.
\newblock Gaussianbody: Clothed human reconstruction via 3d gaussian splatting.
\newblock \emph{arXiv preprint arXiv:2401.09720}, 2024{\natexlab{a}}.

\bibitem[Li et~al.(2022)Li, Tanke, Vo, Zollh{\"o}fer, Gall, Kanazawa, and
  Lassner]{li2022tava}
Ruilong Li, Julian Tanke, Minh Vo, Michael Zollh{\"o}fer, J{\"u}rgen Gall,
  Angjoo Kanazawa, and Christoph Lassner.
\newblock Tava: Template-free animatable volumetric actors.
\newblock In \emph{European Conference on Computer Vision}, pages 419--436.
  Springer, 2022.

\bibitem[Li et~al.(2024{\natexlab{b}})Li, Chen, Li, and Xu]{li2023spacetime}
Zhan Li, Zhang Chen, Zhong Li, and Yi Xu.
\newblock Spacetime gaussian feature splatting for real-time dynamic view
  synthesis.
\newblock In \emph{Proceedings of the IEEE/CVF Conference on Computer Vision
  and Pattern Recognition}, pages 8508--8520, 2024{\natexlab{b}}.

\bibitem[Li et~al.(2024{\natexlab{c}})Li, Zheng, Wang, and
  Liu]{li2024animatable}
Zhe Li, Zerong Zheng, Lizhen Wang, and Yebin Liu.
\newblock Animatable gaussians: Learning pose-dependent gaussian maps for
  high-fidelity human avatar modeling.
\newblock In \emph{Proceedings of the IEEE/CVF Conference on Computer Vision
  and Pattern Recognition}, pages 19711--19722, 2024{\natexlab{c}}.

\bibitem[Li et~al.(2024{\natexlab{d}})Li, Zheng, Wang, and
  Liu]{li2024animatablegaussians}
Zhe Li, Zerong Zheng, Lizhen Wang, and Yebin Liu.
\newblock Animatable gaussians: Learning pose-dependent gaussian maps for
  high-fidelity human avatar modeling.
\newblock In \emph{Proceedings of the IEEE/CVF Conference on Computer Vision
  and Pattern Recognition (CVPR)}, 2024{\natexlab{d}}.

\bibitem[Liang et~al.(2023)Liang, Khan, Li, Nguyen-Phuoc, Lanman, Tompkin, and
  Xiao]{liang2023gaufre}
Yiqing Liang, Numair Khan, Zhengqin Li, Thu Nguyen-Phuoc, Douglas Lanman, James
  Tompkin, and Lei Xiao.
\newblock Gaufre: Gaussian deformation fields for real-time dynamic novel view
  synthesis.
\newblock \emph{arXiv preprint arXiv:2312.11458}, 2023.

\bibitem[Lin et~al.(2021)Lin, Ryabtsev, Sengupta, Curless, Seitz, and
  Kemelmacher-Shlizerman]{lin2021real}
Shanchuan Lin, Andrey Ryabtsev, Soumyadip Sengupta, Brian~L Curless, Steven~M
  Seitz, and Ira Kemelmacher-Shlizerman.
\newblock Real-time high-resolution background matting.
\newblock In \emph{Proceedings of the IEEE/CVF Conference on Computer Vision
  and Pattern Recognition}, pages 8762--8771, 2021.

\bibitem[Lin et~al.(2024)Lin, Dai, Zhu, and Yao]{lin2024gaussian}
Youtian Lin, Zuozhuo Dai, Siyu Zhu, and Yao Yao.
\newblock Gaussian-flow: 4d reconstruction with dynamic 3d gaussian particle.
\newblock In \emph{Proceedings of the IEEE/CVF Conference on Computer Vision
  and Pattern Recognition}, pages 21136--21145, 2024.

\bibitem[Liu et~al.(2024)Liu, Zeng, Ren, Li, Zhang, Yang, Jiang, Li, Yang, Su,
  et~al.]{liu2023grounding}
Shilong Liu, Zhaoyang Zeng, Tianhe Ren, Feng Li, Hao Zhang, Jie Yang, Qing
  Jiang, Chunyuan Li, Jianwei Yang, Hang Su, et~al.
\newblock Grounding dino: Marrying dino with grounded pre-training for open-set
  object detection.
\newblock In \emph{European Conference on Computer Vision}, pages 38--55.
  Springer, 2024.

\bibitem[Loper et~al.(2015)Loper, Mahmood, Romero, Pons-Moll, and
  Black]{SMPL2015}
Matthew Loper, Naureen Mahmood, Javier Romero, Gerard Pons-Moll, and Michael~J.
  Black.
\newblock Smpl: A skinned multi-person linear model.
\newblock \emph{ACM Trans. Graph.}, 34\penalty0 (6):\penalty0 248:1--248:16,
  2015.

\bibitem[Luiten et~al.(2024)Luiten, Kopanas, Leibe, and
  Ramanan]{luiten2023dynamic}
Jonathon Luiten, Georgios Kopanas, Bastian Leibe, and Deva Ramanan.
\newblock Dynamic 3d gaussians: Tracking by persistent dynamic view synthesis.
\newblock In \emph{3DV}, 2024.

\bibitem[Luo et~al.(2022)Luo, Xu, Jiang, Zhou, Qiu, Zhang, Yang, Xu, and
  Yu]{luo2022artemis}
Haimin Luo, Teng Xu, Yuheng Jiang, Chenglin Zhou, Qiwei Qiu, Yingliang Zhang,
  Wei Yang, Lan Xu, and Jingyi Yu.
\newblock Artemis: Articulated neural pets with appearance and motion
  synthesis.
\newblock \emph{ACM Trans. Graph.}, 41\penalty0 (4), 2022.

\bibitem[Medeiros(2024)]{lang-segment-anything}
Luca Medeiros.
\newblock Lang-segment-anything.
\newblock \url{https://github.com/luca-medeiros/lang-segment-anything}, 2024.
\newblock Accessed: 2024-11-08.

\bibitem[Mildenhall et~al.(2020)Mildenhall, Srinivasan, Tancik, Barron,
  Ramamoorthi, and Ng]{nerf}
Ben Mildenhall, Pratul~P. Srinivasan, Matthew Tancik, Jonathan~T. Barron, Ravi
  Ramamoorthi, and Ren Ng.
\newblock Nerf: Representing scenes as neural radiance fields for view
  synthesis.
\newblock In \emph{Computer Vision -- ECCV 2020}, pages 405--421, Cham, 2020.
  Springer International Publishing.

\bibitem[Morton(1966)]{morton1966computer}
Guy~M Morton.
\newblock A computer oriented geodetic data base and a new technique in file
  sequencing.
\newblock 1966.

\bibitem[M\"uller et~al.(2022)M\"uller, Evans, Schied, and
  Keller]{muller2022instant}
Thomas M\"uller, Alex Evans, Christoph Schied, and Alexander Keller.
\newblock Instant neural graphics primitives with a multiresolution hash
  encoding.
\newblock \emph{ACM Trans. Graph.}, 41\penalty0 (4):\penalty0 102:1--102:15,
  2022.

\bibitem[Newcombe et~al.(2015)Newcombe, Fox, and
  Seitz]{newcombe2015dynamicfusion}
Richard~A Newcombe, Dieter Fox, and Steven~M Seitz.
\newblock Dynamicfusion: Reconstruction and tracking of non-rigid scenes in
  real-time.
\newblock In \emph{Proceedings of the IEEE conference on computer vision and
  pattern recognition}, pages 343--352, 2015.

\bibitem[Noguchi et~al.(2021)Noguchi, Sun, Lin, and Harada]{noguchi2021neural}
Atsuhiro Noguchi, Xiao Sun, Stephen Lin, and Tatsuya Harada.
\newblock Neural articulated radiance field.
\newblock In \emph{Proceedings of the IEEE/CVF International Conference on
  Computer Vision}, pages 5762--5772, 2021.

\bibitem[Pang et~al.(2024)Pang, Zhu, Kortylewski, Theobalt, and
  Habermann]{pang2024ash}
Haokai Pang, Heming Zhu, Adam Kortylewski, Christian Theobalt, and Marc
  Habermann.
\newblock Ash: Animatable gaussian splats for efficient and photoreal human
  rendering.
\newblock In \emph{Proceedings of the IEEE/CVF Conference on Computer Vision
  and Pattern Recognition}, pages 1165--1175, 2024.

\bibitem[Park et~al.(2021)Park, Sinha, Barron, Bouaziz, Goldman, Seitz, and
  Martin-Brualla]{park2021nerfies}
Keunhong Park, Utkarsh Sinha, Jonathan~T Barron, Sofien Bouaziz, Dan~B Goldman,
  Steven~M Seitz, and Ricardo Martin-Brualla.
\newblock Nerfies: Deformable neural radiance fields.
\newblock In \emph{Proceedings of the IEEE/CVF International Conference on
  Computer Vision}, pages 5865--5874, 2021.

\bibitem[Qian et~al.(2024{\natexlab{a}})Qian, Kirschstein, Schoneveld, Davoli,
  Giebenhain, and Nie{\ss}ner]{qian2023gaussianavatars}
Shenhan Qian, Tobias Kirschstein, Liam Schoneveld, Davide Davoli, Simon
  Giebenhain, and Matthias Nie{\ss}ner.
\newblock Gaussianavatars: Photorealistic head avatars with rigged 3d
  gaussians.
\newblock In \emph{Proceedings of the IEEE/CVF Conference on Computer Vision
  and Pattern Recognition}, pages 20299--20309, 2024{\natexlab{a}}.

\bibitem[Qian et~al.(2024{\natexlab{b}})Qian, Wang, Mihajlovic, Geiger, and
  Tang]{qian20243dgs}
Zhiyin Qian, Shaofei Wang, Marko Mihajlovic, Andreas Geiger, and Siyu Tang.
\newblock 3dgs-avatar: Animatable avatars via deformable 3d gaussian splatting.
\newblock In \emph{Proceedings of the IEEE/CVF Conference on Computer Vision
  and Pattern Recognition}, pages 5020--5030, 2024{\natexlab{b}}.

\bibitem[Ravi et~al.(2024)Ravi, Gabeur, Hu, Hu, Ryali, Ma, Khedr, R{\"a}dle,
  Rolland, Gustafson, et~al.]{ravi2024sam}
Nikhila Ravi, Valentin Gabeur, Yuan-Ting Hu, Ronghang Hu, Chaitanya Ryali,
  Tengyu Ma, Haitham Khedr, Roman R{\"a}dle, Chloe Rolland, Laura Gustafson,
  et~al.
\newblock Sam 2: Segment anything in images and videos.
\newblock \emph{arXiv preprint arXiv:2408.00714}, 2024.

\bibitem[Rong et~al.(2024)Rong, Grigorev, Wang, Black, Thomaszewski,
  Tsalicoglou, and Hilliges]{rong2024gaussian}
Boxiang Rong, Artur Grigorev, Wenbo Wang, Michael~J Black, Bernhard
  Thomaszewski, Christina Tsalicoglou, and Otmar Hilliges.
\newblock Gaussian garments: Reconstructing simulation-ready clothing with
  photorealistic appearance from multi-view video.
\newblock \emph{arXiv preprint arXiv:2409.08189}, 2024.

\bibitem[Ronneberger et~al.(2015)Ronneberger, Fischer, and
  Brox]{ronneberger2015u}
Olaf Ronneberger, Philipp Fischer, and Thomas Brox.
\newblock U-net: Convolutional networks for biomedical image segmentation.
\newblock In \emph{International Conference on Medical image computing and
  computer-assisted intervention}, pages 234--241. Springer, 2015.

\bibitem[Shao et~al.(2024)Shao, Wang, Tian, Yang, Meng, Cai, Dong, Zhang,
  Zhang, and Wang]{shao2024degas}
Zhijing Shao, Duotun Wang, Qing-Yao Tian, Yao-Dong Yang, Hengyu Meng, Zeyu Cai,
  Bo Dong, Yu Zhang, Kang Zhang, and Zeyu Wang.
\newblock Degas: Detailed expressions on full-body gaussian avatars.
\newblock \emph{arXiv preprint arXiv:2408.10588}, 2024.

\bibitem[Shen et~al.(2023)Shen, Guo, Kaufmann, Zarate, Valentin, Song, and
  Hilliges]{shen2023x}
Kaiyue Shen, Chen Guo, Manuel Kaufmann, Juan~Jose Zarate, Julien Valentin, Jie
  Song, and Otmar Hilliges.
\newblock X-avatar: Expressive human avatars.
\newblock In \emph{Proceedings of the IEEE/CVF Conference on Computer Vision
  and Pattern Recognition}, pages 16911--16921, 2023.

\bibitem[Shysheya et~al.(2019)Shysheya, Zakharov, Aliev, Bashirov, Burkov,
  Iskakov, Ivakhnenko, Malkov, Pasechnik, Ulyanov,
  et~al.]{shysheya2019textured}
Aliaksandra Shysheya, Egor Zakharov, Kara-Ali Aliev, Renat Bashirov, Egor
  Burkov, Karim Iskakov, Aleksei Ivakhnenko, Yury Malkov, Igor Pasechnik,
  Dmitry Ulyanov, et~al.
\newblock Textured neural avatars.
\newblock In \emph{Proceedings of the IEEE/CVF Conference on Computer Vision
  and Pattern Recognition}, pages 2387--2397, 2019.

\bibitem[Slavcheva et~al.(2017)Slavcheva, Baust, Cremers, and
  Ilic]{KillingFusion2017cvpr}
Miroslava Slavcheva, Maximilian Baust, Daniel Cremers, and Slobodan Ilic.
\newblock Killingfusion: Non-rigid 3d reconstruction without correspondences.
\newblock In \emph{Proceedings of the IEEE Conference on Computer Vision and
  Pattern Recognition}, pages 1386--1395, 2017.

\bibitem[Slavcheva et~al.(2018)Slavcheva, Baust, and
  Ilic]{slavcheva2018sobolevfusion}
Miroslava Slavcheva, Maximilian Baust, and Slobodan Ilic.
\newblock Sobolevfusion: 3d reconstruction of scenes undergoing free non-rigid
  motion.
\newblock In \emph{Proceedings of the IEEE conference on computer vision and
  pattern recognition}, pages 2646--2655, 2018.

\bibitem[Song et~al.(2023)Song, Chen, Li, Chen, Chen, Yuan, Xu, and
  Geiger]{song2023nerfplayer}
Liangchen Song, Anpei Chen, Zhong Li, Zhang Chen, Lele Chen, Junsong Yuan, Yi
  Xu, and Andreas Geiger.
\newblock Nerfplayer: A streamable dynamic scene representation with decomposed
  neural radiance fields.
\newblock \emph{IEEE Transactions on Visualization and Computer Graphics},
  29\penalty0 (5):\penalty0 2732--2742, 2023.

\bibitem[Stoll et~al.(2011)Stoll, Hasler, Gall, Seidel, and
  Theobalt]{StollHGST2011}
Carsten Stoll, Nils Hasler, Juergen Gall, Hans-Peter Seidel, and Christian
  Theobalt.
\newblock Fast articulated motion tracking using a sums of {Gaussians} body
  model.
\newblock In \emph{International Conference on Computer Vision (ICCV)}, 2011.

\bibitem[Su et~al.(2020)Su, Xu, Zheng, Yu, Liu, and Fang]{robustfusion}
Zhuo Su, Lan Xu, Zerong Zheng, Tao Yu, Yebin Liu, and Lu Fang.
\newblock Robustfusion: Human volumetric capture with data-driven visual cues
  using a rgbd camera.
\newblock In \emph{Computer Vision -- ECCV 2020}, pages 246--264, Cham, 2020.
  Springer International Publishing.

\bibitem[Su et~al.(2022)Su, Xu, Zhong, Li, Deng, Quan, and
  Fang]{su2022robustfusionPlus}
Zhuo Su, Lan Xu, Dawei Zhong, Zhong Li, Fan Deng, Shuxue Quan, and Lu Fang.
\newblock Robustfusion: Robust volumetric performance reconstruction under
  human-object interactions from monocular rgbd stream.
\newblock \emph{IEEE Transactions on Pattern Analysis and Machine
  Intelligence}, 2022.

\bibitem[Sumner and Popovi{\'c}(2004)]{sumner2004deformation}
Robert~W Sumner and Jovan Popovi{\'c}.
\newblock Deformation transfer for triangle meshes.
\newblock \emph{ACM Transactions on graphics (TOG)}, 23\penalty0 (3):\penalty0
  399--405, 2004.

\bibitem[Sumner et~al.(2007)Sumner, Schmid, and Pauly]{sumner2007embedded}
Robert~W Sumner, Johannes Schmid, and Mark Pauly.
\newblock Embedded deformation for shape manipulation.
\newblock \emph{ACM Transactions on Graphics (TOG)}, 26\penalty0 (3):\penalty0
  80, 2007.

\bibitem[Sun et~al.(2021)Sun, Chen, Chen, Pang, Lin, Jiang, Xu, Wang, and
  Yu]{sun2021HOI-FVV}
Guoxing Sun, Xin Chen, Yizhang Chen, Anqi Pang, Pei Lin, Yuheng Jiang, Lan Xu,
  Jingya Wang, and Jingyi Yu.
\newblock Neural free-viewpoint performance rendering under complex
  human-object interactions.
\newblock In \emph{Proceedings of the 29th ACM International Conference on
  Multimedia}, 2021.

\bibitem[Suo et~al.(2021)Suo, Jiang, Lin, Zhang, Wu, Guo, and
  Xu]{suo2021neuralhumanfvv}
Xin Suo, Yuheng Jiang, Pei Lin, Yingliang Zhang, Minye Wu, Kaiwen Guo, and Lan
  Xu.
\newblock Neuralhumanfvv: Real-time neural volumetric human performance
  rendering using rgb cameras.
\newblock In \emph{Proceedings of the IEEE/CVF Conference on Computer Vision
  and Pattern Recognition}, pages 6226--6237, 2021.

\bibitem[Tretschk et~al.(2021)Tretschk, Tewari, Golyanik, Zollh\"{o}fer,
  Lassner, and Theobalt]{tretschk2021nonrigid}
Edgar Tretschk, Ayush Tewari, Vladislav Golyanik, Michael Zollh\"{o}fer,
  Christoph Lassner, and Christian Theobalt.
\newblock Non-rigid neural radiance fields: Reconstruction and novel view
  synthesis of a dynamic scene from monocular video.
\newblock In \emph{{IEEE} International Conference on Computer Vision
  ({ICCV})}. {IEEE}, 2021.

\bibitem[Uzolas et~al.(2024)Uzolas, Eisemann, and
  Kellnhofer]{uzolas2024template}
Lukas Uzolas, Elmar Eisemann, and Petr Kellnhofer.
\newblock Template-free articulated neural point clouds for reposable view
  synthesis.
\newblock \emph{Advances in Neural Information Processing Systems}, 36, 2024.

\bibitem[Wang et~al.(2024{\natexlab{a}})Wang, Yao, Guo, Zhang, Hu, Yu, Xu, and
  Wu]{wang2023videorf}
Liao Wang, Kaixin Yao, Chengcheng Guo, Zhirui Zhang, Qiang Hu, Jingyi Yu, Lan
  Xu, and Minye Wu.
\newblock Videorf: Rendering dynamic radiance fields as 2d feature video
  streams.
\newblock In \emph{Proceedings of the IEEE/CVF Conference on Computer Vision
  and Pattern Recognition}, pages 470--481, 2024{\natexlab{a}}.

\bibitem[Wang et~al.(2024{\natexlab{b}})Wang, Zhang, Wang, Yao, Xie, Yu, Wu,
  and Xu]{wang2024v}
Penghao Wang, Zhirui Zhang, Liao Wang, Kaixin Yao, Siyuan Xie, Jingyi Yu, Minye
  Wu, and Lan Xu.
\newblock V\^{} 3: Viewing volumetric videos on mobiles via streamable 2d
  dynamic gaussians.
\newblock \emph{ACM Transactions on Graphics (TOG)}, 43\penalty0 (6):\penalty0
  1--13, 2024{\natexlab{b}}.

\bibitem[Wang et~al.(2022)Wang, Schwarz, Geiger, and Tang]{ARAH:ECCV:2022}
Shaofei Wang, Katja Schwarz, Andreas Geiger, and Siyu Tang.
\newblock Arah: Animatable volume rendering of articulated human sdfs.
\newblock In \emph{European Conference on Computer Vision}, 2022.

\bibitem[Wang et~al.(2023)Wang, Han, Habermann, Daniilidis, Theobalt, and
  Liu]{wang2023neus2}
Yiming Wang, Qin Han, Marc Habermann, Kostas Daniilidis, Christian Theobalt,
  and Lingjie Liu.
\newblock Neus2: Fast learning of neural implicit surfaces for multi-view
  reconstruction.
\newblock In \emph{Proceedings of the IEEE/CVF International Conference on
  Computer Vision}, pages 3295--3306, 2023.

\bibitem[Weng et~al.(2022)Weng, Curless, Srinivasan, Barron, and
  Kemelmacher-Shlizerman]{weng_humannerf_2022_cvpr}
Chung-Yi Weng, Brian Curless, Pratul~P. Srinivasan, Jonathan~T. Barron, and Ira
  Kemelmacher-Shlizerman.
\newblock Human{N}e{RF}: Free-viewpoint rendering of moving people from
  monocular video.
\newblock In \emph{Proceedings of the IEEE/CVF Conference on Computer Vision
  and Pattern Recognition (CVPR)}, pages 16210--16220, 2022.

\bibitem[Xiang et~al.(2021)Xiang, Prada, Bagautdinov, Xu, Dong, Wen, Hodgins,
  and Wu]{xiang2021modeling}
Donglai Xiang, Fabian Prada, Timur Bagautdinov, Weipeng Xu, Yuan Dong, He Wen,
  Jessica Hodgins, and Chenglei Wu.
\newblock Modeling clothing as a separate layer for an animatable human avatar.
\newblock \emph{ACM Transactions on Graphics (TOG)}, 40\penalty0 (6):\penalty0
  1--15, 2021.

\bibitem[Xiang et~al.(2022)Xiang, Bagautdinov, Stuyck, Prada, Romero, Xu,
  Saito, Guo, Smith, Shiratori, Sheikh, Hodgins, and Wu]{xiang2022dressing}
Donglai Xiang, Timur Bagautdinov, Tuur Stuyck, Fabian Prada, Javier Romero,
  Weipeng Xu, Shunsuke Saito, Jingfan Guo, Breannan Smith, Takaaki Shiratori,
  Yaser Sheikh, Jessica Hodgins, and Chenglei Wu.
\newblock Dressing avatars: Deep photorealistic appearance for physically
  simulated clothing.
\newblock \emph{ACM Trans. Graph.}, 41\penalty0 (6), 2022.

\bibitem[Xu et~al.(2019)Xu, Su, Han, Yu, Liu, and Fang]{UnstructureLan}
Lan Xu, Zhuo Su, Lei Han, Tao Yu, Yebin Liu, and Lu Fang.
\newblock Unstructuredfusion: realtime 4d geometry and texture reconstruction
  using commercial rgbd cameras.
\newblock \emph{IEEE transactions on pattern analysis and machine
  intelligence}, 42\penalty0 (10):\penalty0 2508--2522, 2019.

\bibitem[Xu et~al.(2018)Xu, Chatterjee, Zollh\"{o}fer, Rhodin, Mehta, Seidel,
  and Theobalt]{MonoPerfCap}
Weipeng Xu, Avishek Chatterjee, Michael Zollh\"{o}fer, Helge Rhodin, Dushyant
  Mehta, Hans-Peter Seidel, and Christian Theobalt.
\newblock Monoperfcap: Human performance capture from monocular video.
\newblock \emph{ACM Transactions on Graphics (TOG)}, 37\penalty0 (2):\penalty0
  27:1--27:15, 2018.

\bibitem[Yu et~al.(2021)Yu, Li, Tancik, Li, Ng, and
  Kanazawa]{yu2021plenoctrees}
Alex Yu, Ruilong Li, Matthew Tancik, Hao Li, Ren Ng, and Angjoo Kanazawa.
\newblock Plenoctrees for real-time rendering of neural radiance fields.
\newblock In \emph{Proceedings of the IEEE/CVF International Conference on
  Computer Vision}, pages 5752--5761, 2021.

\bibitem[Zhao et~al.(2022)Zhao, Jiang, Yao, Zhang, Wang, Dai, Zhong, Zhang, Wu,
  Xu, and Yu]{zhao2022human}
Fuqiang Zhao, Yuheng Jiang, Kaixin Yao, Jiakai Zhang, Liao Wang, Haizhao Dai,
  Yuhui Zhong, Yingliang Zhang, Minye Wu, Lan Xu, and Jingyi Yu.
\newblock Human performance modeling and rendering via neural animated mesh.
\newblock \emph{ACM Trans. Graph.}, 41\penalty0 (6), 2022.

\bibitem[Zheng et~al.(2024{\natexlab{a}})Zheng, Wen, Li, Zhang, Su, Chang,
  Zhao, Lv, Zhang, Zhang, et~al.]{zheng2024headgap}
Xiaozheng Zheng, Chao Wen, Zhaohu Li, Weiyi Zhang, Zhuo Su, Xu Chang, Yang
  Zhao, Zheng Lv, Xiaoyuan Zhang, Yongjie Zhang, et~al.
\newblock Headgap: Few-shot 3d head avatar via generalizable gaussian priors.
\newblock \emph{arXiv preprint arXiv:2408.06019}, 2024{\natexlab{a}}.

\bibitem[Zheng et~al.(2024{\natexlab{b}})Zheng, Wen, Su, Xu, Li, Zhao, and
  Xue]{zheng2024ohta}
Xiaozheng Zheng, Chao Wen, Zhuo Su, Zeran Xu, Zhaohu Li, Yang Zhao, and Zhou
  Xue.
\newblock Ohta: One-shot hand avatar via data-driven implicit priors.
\newblock In \emph{Proceedings of the IEEE/CVF Conference on Computer Vision
  and Pattern Recognition}, pages 799--810, 2024{\natexlab{b}}.

\bibitem[Zheng et~al.(2023)Zheng, Zhao, Zhang, Liu, and
  Liu]{zheng2023avatarrex}
Zerong Zheng, Xiaochen Zhao, Hongwen Zhang, Boning Liu, and Yebin Liu.
\newblock Avatarrex: Real-time expressive full-body avatars.
\newblock \emph{ACM Transactions on Graphics (TOG)}, 42\penalty0 (4), 2023.

\bibitem[Zhu et~al.(2024)Zhu, Zhan, Theobalt, and Habermann]{zhu2023trihuman}
Heming Zhu, Fangneng Zhan, Christian Theobalt, and Marc Habermann.
\newblock Trihuman: a real-time and controllable tri-plane representation for
  detailed human geometry and appearance synthesis.
\newblock \emph{ACM Transactions on Graphics}, 44\penalty0 (1):\penalty0 1--17,
  2024.

\bibitem[Zielonka et~al.(2025)Zielonka, Bagautdinov, Saito, Zollhöfer, Thies,
  and Romero]{zielonka25dega}
Wojciech Zielonka, Timur Bagautdinov, Shunsuke Saito, Michael Zollhöfer,
  Justus Thies, and Javier Romero.
\newblock Drivable 3d gaussian avatars.
\newblock In \emph{International Conference on 3D Vision (3DV)}, 2025.

\end{thebibliography}
}

\appendix
\clearpage
\setcounter{page}{1}
\maketitlesupplementary

Within the supplementary material, we provide:

\begin{itemize}
\item Implementation details in Appendix Sec.~\ref{suppl:detail}.
\item Qualitative and Quantitative comparison with Animatable Gaussians~\cite{li2024animatable} in Appendix Sec.~\ref{suppl:comparison}.
\item Additional ablative studies in Appendix Sec.~\ref{suppl:ablation}.
\end{itemize}

\section{Implementation Details} \label{suppl:detail} 

\noindent{\bf Initialization.} We initialize approximately 50,000 motion Gaussians in the canonical frame using a uniform random distribution. During subsequent optimization, aside from the photometric loss in Eq.~\ref{eq:1},
the isotropic loss and size loss are defined as follows:
\begin{equation}
\begin{split}
E_{\mathrm{iso}} & =\frac{1}{N} \sum_{i=1}^{N} \text{ReLU}(e^{max(s_i) - min(s_i)} - r), \\
E_{\text {size }} &=\sum_{i=1}^N \text{ReLU} \left(s_i- \alpha \frac{1}{N} \sum_{i=1}^N sg[s_i] \right), 
\end{split}
\end{equation}

where $s_i$ represents the scaling parameters of the $i$-th Gaussian, $N$ denotes the number of Gaussians, and $e$ is the exponential activation function. The isotropic loss ensures the ratio between the major and minor axes of each Gaussian does not exceed $r$ (set to 4 in our experiments). The $sg$ denotes stop-gradient operator.  
For appearance Gaussians, we use the initialized motion Gaussians as input and further optimize it with densification and prune as the original 3DGS~\cite{kerbl20233d}. 

\begin{figure}[t] 
	\begin{center} 
		\includegraphics[width=\linewidth]{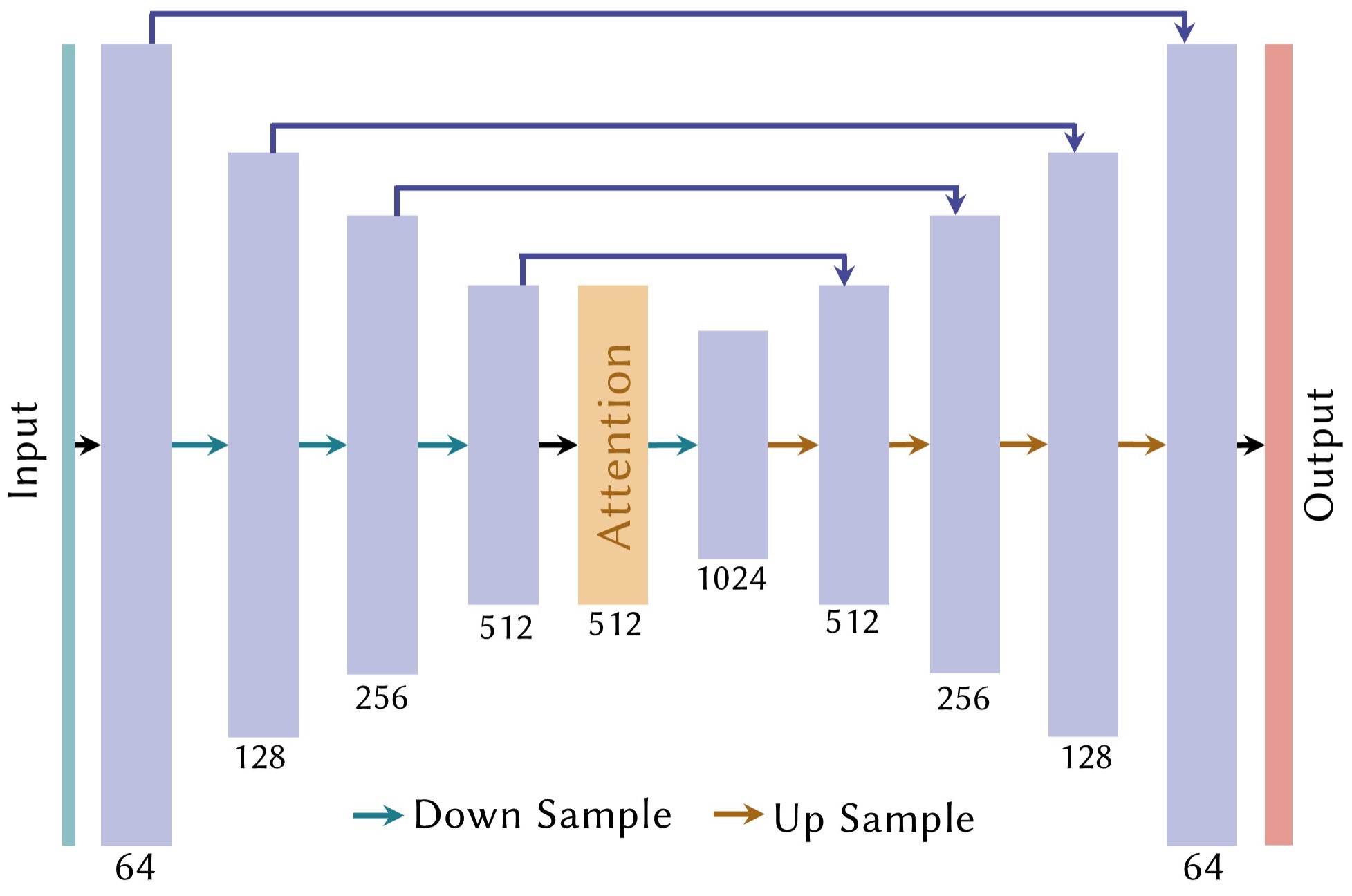} 
	\end{center} 
    \vspace{-20pt}  
	\caption{Network architecture.}
	\label{fig:network} 
	\vspace{-5pt}
\end{figure}

\noindent{\bf Network Architecture.} We use three identical U-Net architectures, adjusting the output feature dimensions to accommodate different attributes. Each U-Net incorporates a self-attention layer to maintain global consistency. As illustrated in Fig.~\ref{fig:network}, the self-attention layer is applied before the final downsampling step of the U-Net.

\begin{figure}[htbp] 
	\begin{center} 
		\includegraphics[width=0.99\linewidth]{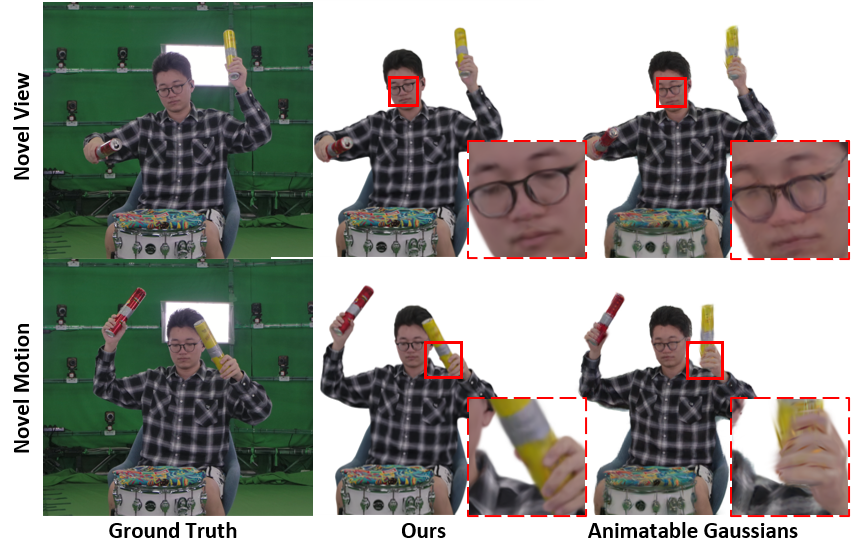} 
	\end{center} 
	\vspace{-16pt}
	\caption{Qualitative comparison with Animatable Gaussians~\cite{li2024animatable}.}
	\label{fig:ag}
\end{figure}

\noindent{\bf Re-performance.} During the alignment phase, we assign semantic labels to specific regions, including the head, left hand, right hand, left foot, and right foot, to align corresponding clusters between the source and target Gaussians. For objects, we use pairwise semantic prompts (e.g., ``basketball" and ``balloon") and assign consistent labels during the unprojection process, ensuring accurate alignment across different objects.
The re-performance stage employs the Adam optimizer, with the alignment phase trained for 15,000 iterations and the motion transfer phase for 2,000 iterations.

\begin{figure*}[htbp] 
	\begin{center} 
		\includegraphics[width=0.99\linewidth]{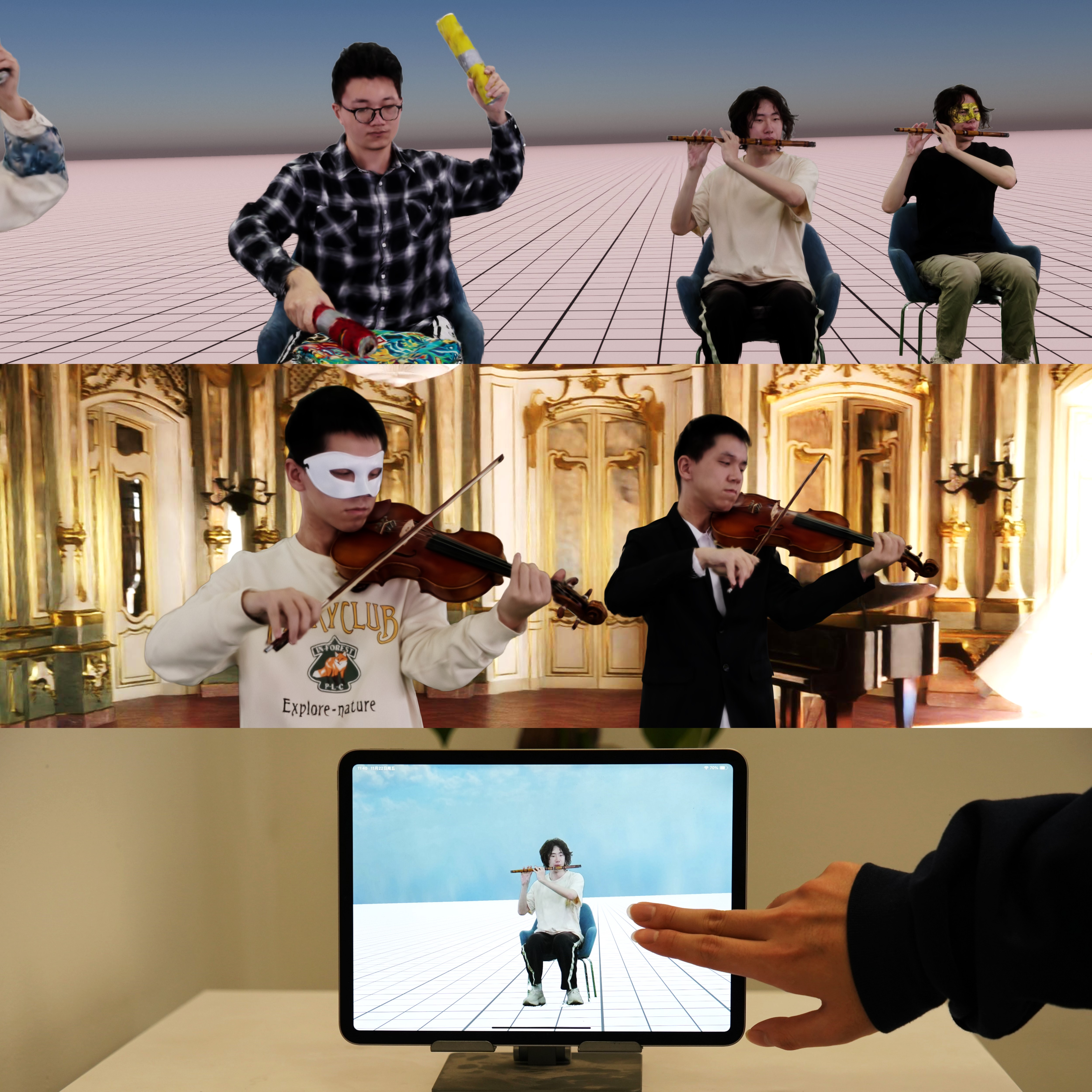} 
	\end{center} 
	\vspace{-16pt}
	\caption{We further compress our Gaussian sequences using DualGS, demonstrating their adaptability in various immersive applications.}
	\label{fig:app}
\end{figure*}

Our method efficiently generates Gaussian sequences for high-fidelity playback and vivid re-performance of general non-rigid scenes. We train the model using the PyTorch framework on a single NVIDIA GeForce RTX3090 GPU, achieving a rendering speed of 7 FPS. The Gaussian sequences can be further baked and compressed via DualGS~\cite{jiang2024robust}, allowing seamless integration into low-end devices like VR headsets and iPads for intuitive, user-friendly interaction as demonstrated in Fig.~\ref{fig:app}.

\section{Comparison} \label{suppl:comparison} 

We further compare our method with Animatable Gaussians~\cite{li2024animatable} to evaluate its rendering quality on novel views and motions. As shown in Fig.~\ref{fig:ag} and Tab.~\ref{table:ag}, this method relies heavily on the human parametric model SMPL~\cite{SMPL2015}, which constrains its ability to accurately estimate motion in regions far from the human body. This limitation results in severe artifacts when handling general non-rigid scenarios.

\begin{table}[t]
    \centering

    \resizebox{\columnwidth}{!}{ %
    \begin{tabular}{c|c c c|c c c}
        \hline
        \multirow{2}{*}{\textbf{Methods}} & \multicolumn{3}{c|}{\textbf{Novel View}} & \multicolumn{3}{c}{\textbf{Novel Motion}} \\
        & PSNR $\uparrow$ & SSIM $\uparrow$  &  LPIPS $\downarrow$ & PSNR $\uparrow$ & SSIM $\uparrow$  &  LPIPS $\downarrow$ \\
        \hline
        \hline %
        Animatable Gaussians  & 24.26 & 0.934 & 0.0647 & 22.04 & 0.911 & 0.079 \\
             \hline

        Ours  & 32.09 & 0.979 & 0.0310 & 30.06 & 0.976 & 0.0277 \\
        \hline
    \end{tabular}
    }
    \caption{
    Quantitative comparison with Animatable Gaussians~\cite{li2024animatable}. 
    }
    \label{table:ag}
\end{table}

\begin{table}[t]
    \centering

    \resizebox{\columnwidth}{!}{ %
    \begin{tabular}{l|ccc}
        \hline
        Methods       & PSNR $\uparrow$ & SSIM $\uparrow$  &  LPIPS $\downarrow$\\
        \hline
        w/20.cam  & 30.36 & 0.971 & 0.0471  \\
        w/40.cam   & 30.89 & 0.975 &  0.0447  \\
        w/60.cam   & 31.33 & 0.977 & 0.0429  \\
        \hline %
        Ours  & 32.09 & 0.979 & 0.0310 \\
        \hline
    \end{tabular}
    }
    \caption{ 
    Quantitative Ablation Study on the different input view numbers.
    }
    \label{table:ab1}
\end{table}

\begin{table}[t]
	\begin{center}
		\centering
		\vspace{20pt}

		\resizebox{0.47\textwidth}{!}{
			\begin{tabular}{l|ccc}
				\hline
				Methods   &  PSNR $\uparrow$ & SSIM $\uparrow$ & LPIPS $\downarrow$ \\
				\hline
				w/ 64.res & 28.85 & 0.949 & 0.0887 \\
			    w/ 128.res & 32.82 & 0.981 & 0.0413 \\
                w/ 256.res & 33.45 & 0.986 & 0.0283 \\
                w/o atten  & 34.16 & 0.990 & 0.0197 \\
    		\hline
                \textbf{Ours} & 34.27 & 0.989 & 0.0227 \\
				\hline
			\end{tabular}
            }
            \caption{
            Quantitative Ablation Study on the resolution of position maps.
            }
		\label{table:ab2}
        \vspace{-10pt}
	\end{center}
\end{table}

\section{Ablations} \label{suppl:ablation} 
\noindent{\bf Number of Camera Views.} To assess the robustness of RePerformer under sparser input views, we perform ablation experiments using uniformly selected subsets of 20, 40, and 60 camera views for training, denoted as w/20.cam, w/40.cam, and w/60.cam. As shown in Tab.~\ref{table:ab1}, our method maintains satisfactory rendering quality even with sparser input views.

\noindent{\bf Position Map Resolution.} As illustrated in Tab.~\ref{table:ab2}, we conduct ablation studies with position map resolutions of 64, 128, 256, and 512 (ours), corresponding to different numbers of Gaussians. Although larger position maps can store denser Gaussians, they exceed GPU memory limitations. We also compare results without the self-attention layer. Our full pipeline achieves high-fidelity rendering quality at a resolution of 512 with the self-attention module, as shown in Tab.~\ref{table:ab2}.

\end{document}